%% file: camera-ready_main.tex
\documentclass{article}

\usepackage[preprint]{neurips_2026}

\usepackage[utf8]{inputenc} 
\usepackage[T1]{fontenc}    
\usepackage{hyperref}       
\usepackage{url}            
\usepackage{booktabs}       
\usepackage{amsfonts}       
\usepackage{nicefrac}       
\usepackage{microtype}      
\usepackage{xcolor}         
\usepackage{subcaption}

\usepackage{amsmath}
\usepackage{amssymb}
\usepackage{mathtools}
\usepackage{amsthm}
\usepackage{url}            
\usepackage{booktabs}       
\usepackage{amsfonts,bbm}   
\usepackage{nicefrac}       
\usepackage{microtype}      
\usepackage{xcolor}         

\usepackage{algorithm}
\usepackage{algorithmic}
\usepackage{dsfont}

\usepackage{bbm}
\usepackage{mathtools}
\usepackage[capitalize,noabbrev]{cleveref}

\theoremstyle{plain}
\newtheorem{theorem}{Theorem}[section]
\newtheorem{proposition}[theorem]{Proposition}
\newtheorem{lemma}[theorem]{Lemma}

\theoremstyle{definition}
\newtheorem{definition}[theorem]{Definition}
\newtheorem{assumption}[theorem]{Assumption}
\theoremstyle{remark}
\newtheorem{remark}[theorem]{Remark}

\crefname{theorem}{Theorem}{Theorems}
\Crefname{theorem}{Theorem}{Theorems}

\crefname{lemma}{Lemma}{Lemmas}
\Crefname{lemma}{Lemma}{Lemmas}

\crefname{proposition}{Proposition}{Propositions}
\Crefname{proposition}{Proposition}{Propositions}

\crefname{corollary}{Corollary}{Corollaries}
\Crefname{corollary}{Corollary}{Corollaries}

\crefname{definition}{Definition}{Definitions}
\Crefname{definition}{Definition}{Definitions}

\crefname{remark}{Remark}{Remarks}
\Crefname{remark}{Remark}{Remarks}

\crefname{assumption}{Assumption}{Assumptions}
\Crefname{assumption}{Assumption}{Assumptions}

\newcommand{\ind}{\mathds{1}}

\newcommand{\mbbR}{\mathbb{R}}
\newcommand{\mbbP}{\mathbb{P}}

\newcommand{\mX}{\mathcal{X}}

\newcommand{\mY}{\mathcal{Y}}

\newcommand{\mI}{\mathcal{I}}
\newcommand{\mG}{\mathcal{G}}
\newcommand{\mR}{\mathcal{R}}
\newcommand{\mD}{\mathcal{D}}

\newcommand{\whT}{\widehat{T}}

\newcommand{\NA}{\mathrm{NA}}

\newcommand{\obs}{\mathrm{obs}}
\newcommand{\mis}{\mathrm{mis}}
\newcommand{\tree}{\mathrm{tree}}

\title{Beyond Imputation: Mask-Adaptive Conformal Prediction via Tree Embeddings on General Missing Data Mechanisms}

\author{%
	Jiarong Fan\thanks{\texttt{jiarong.fan@universite-paris-saclay.fr}} \\
	LaMME \\
	Universit\'e Paris-Saclay
	\And
	Juhyun Park \\
	LaMME, ENSIIE \\
	Universit\'e Paris-Saclay
	\AND
	Thi Phuong Thuy Vo \\
	LaMME, ENSIIE\\
	Universit\'e Paris-Saclay
	\And
	Nicolas J-B.~Brunel \\
	LaMME, ENSIIE\\
	Universit\'e Paris-Saclay \\
	Capgemini Invent
}

\begin{document}

\maketitle

\begin{abstract}
   Conformal prediction (CP) provides a distribution-free framework for uncertainty quantification with finite-sample marginal coverage guarantees. Yet, missing values introduce distributional heterogeneity, where marginal coverage can hide systematic undercoverage for certain missingness patterns. At the same time,  existing methods based on distances or smoothness require a continuous representation of partially observed covariates. Although imputation restores continuity, it obscures the missingness patterns and introduces task-dependent overhead. 
   We develop a new CP framework that is \emph{adaptive to missingness patterns}, without imputing covariates. We rely on data augmentation by feature embedding that accounts for missingness. Our framework represents each input with missing values by a tree embedding together with its raw missingness mask, and then learns a regularized quantile score function. Varying the choices for the penalty function and the augmentation enables us to characterize three practically relevant validity levels: exact coordinate-wise validity, exact validity on observed masks, and approximate missingness-conditional guarantees for general mechanisms (MCAR, MAR, MNAR). 
	Experiments on synthetic and real datasets support these guarantees and show that the proposed method reduces miscoverage risk while maintaining informative intervals, especially in non-MCAR settings. These results highlight the practical value of uncertainty quantification that adapts to missingness.
\end{abstract}

\section{Introduction}
Modern machine learning systems are increasingly deployed in safety-critical domains, from medical diagnosis to autonomous driving. In such settings, practitioners cannot rely on point predictions alone; they require \emph{prediction sets}, e.g., intervals in regression or label subsets in classification, that contain the unknown response with a predefined high probability (e.g., 90\%). 
That is, given a sample $(X,Y)\sim \mbbP$ with $X \in \mathcal{X}$ and $Y \in \mathcal{Y}$, and a miscoverage level $\alpha$, one may wish to construct a \emph{marginally} valid set ${C}_\alpha(X)$ such that
$
\mathbb{P}(Y \in {C}_\alpha(X)) \ge 1 - \alpha.
$
Conformal prediction (CP) is a distribution-free uncertainty quantification framework that offers an attractive solution.
Under the exchangeability assumption that the sequence of data points can be permuted without altering their joint distribution, CP wraps any black-box predictor with $\hat C_\alpha$ satisfying finite-sample marginal coverage guarantees. 

However, marginal coverage does not prevent systematic failures on certain subgroups of features. In particular, features with missing values are ubiquitous due to sensor failure, privacy filtering, or cost constraints \citep{emmanuel2021survey}, and can invalidate the adaptivity of coverage \citep{zaffran2023conformal,zaffran2024predictiveuncertaintyquantificationmissing}. Let $M \in \{0,1\}^d$ denote a binary mask indicating which features in $X$ are missing,  and extend the probability measure $\mathbb P$ to the joint $(X,Y,M)$. The presence of missing entries introduces data heterogeneity and can  
undercover when some clinically or operationally important features are absent \citep{zaffran2023conformal}. 
This motivates \emph{mask-conditional coverage} (MCV), which targets
\(
\mathbb P(Y \in \hat C_\alpha(X)\,\big|\, M=m) \geq 1-\alpha,\text{ } \forall m \in \{0,1\}^d,
\)
or weaker but useful variants, such as conditional coverage on the coordinate-wise missingness.
This would matter when certain measurements are more important than others, as observed in clinical settings \cite{singh2021missingness}, where model performance varies across subgroups.
Unfortunately, as in the classical impossibility results for $X$-conditional coverage \citep{lei2014distribution}, achieving exact mask-conditional validity for all patterns can make prediction sets overly conservative (uninformative) \citep{zaffran2024predictiveuncertaintyquantificationmissing}. 
This motivates the central question of this paper:
\emph{How to develop a robust pipeline to construct valid and informative prediction sets, conditional on the missingness patterns, under general mechanisms?}

Classical methods typically rely on impute-then-predict pipelines, followed by standard uncertainty quantification tools. This can preserve marginal coverage, but it obscures the missingness pattern $M$ and does not ensure mask-conditional coverage. Recent work has proposed a missing data augmentation (MDA) technique that modifies either the calibration or the test point, ensuring that all points share the same mask and achieving MCV. The available theoretical guarantees are limited to Missing Completely at Random (MCAR; i.e., $M \perp (X,Y)$) \citep{zaffran2023conformal}. 
For more general missingness,  \citet{fan2025arxiv} proposes a preimpute-then-mask framework with a density-ratio correction, whose performance depends on the quality of the ratio estimator. 
Meanwhile, \citet{gibbs2025conformal, kaur2025conformal} propose relaxations of the conditional coverage. This, however, depends on the smoothness of $X\mapsto Y$ (e.g., via RKHS or Lipschitz models), and is not compatible when the space with missingness lacks a meaningful geometry. Otherwise, it requires choosing an appropriate feature representation and target directions.
A detailed summary of related work is in \cref{appendix:relatedwork}. 

\paragraph{Our contribution.} 

The key idea behind our approach is to create a mask-adaptive feature embedding.
We exploit the capability of tree ensembles to naturally handle masked values via tree splits, producing embedding features for $X$ in a high-dimensional space.
We augment this embedding with purposefully designed mask vectors. A pinball loss function with a carefully designed penalty is then used to learn the quantile score on the augmented embedding space. Inspired by \citet{gibbs2025conformal}, we specifically design augmentation methods and penalties corresponding to three validity criteria of practical relevance under missingness.

Notably, we avoid explicit imputation. While imputation is widely used in practice, it incurs a high computational cost, is task–dependent, and destroys mask information that is important for adaptive uncertainty quantification. Although distributional and multiple imputation can be useful across downstream tasks \citep{naf2024good}, every downstream application could require different imputation functions. We do not claim that imputation is intrinsically unsuitable, but our approach instead avoids additional imputations and preserves the masked inputs.
Our main contributions are as follows:

(1) \textbf{Mask adaptive conformal prediction.} We introduce a general framework for CP with partially observed covariates. We exploit tree-based models to define a deterministic embedding, conditional on the training split, that is adaptive to missingness patterns. 

(2) \textbf{Theoretical guarantees under general missingness.} 
For each validity target of interest, we identify the conditions with which the guarantees can be derived.
In particular, we show that our methods achieve finite sample coverage conditional on individual mask coordinates, i.e., $\mathbb P(Y \in \hat C_\alpha(X)\mid M_j=1)\ge 1-\alpha$, and we provide approximate mask-conditional guarantees for all patterns $m$ with non-asymptotic bounds. 

(3) \textbf{Empirical validation on synthetic and real data.} 
We evaluate the method on synthetic data with controlled MCAR/MAR/MNAR mechanisms, a semi-synthetic Concrete benchmark, and a high-dimensional MIMIC-IV length-of-stay task.
The experiments show favorable coverage-length trade-offs in high-dimensional and mask-heterogeneous settings. Code is available at \url{https://github.com/jrfan/mask_conditional_coverage}.

\section{Background}
Suppose we have $n$ random samples $(X^i, Y^i, M^i) \in \mX \times \mY \times \{0,1\}^d,\,  i=1,\dots,n$, where $\mX= \prod_{j=1}^d \bigl(\mX_j^o  \cup \{\NA\}\bigr) $ is a $d$-dimensional space of mixed-type variables (continuous, discrete, or missing denoted by $\NA$). The ``observable'' space $\mX_j^o$ may differ across components $j=1,\dots,d$. We use the convention $M_j=1$ when coordinate $j$ is missing and $M_j=0$ otherwise. Throughout, we let $\mathsf m\colon\mathcal X\to\{0,1\}^d$ be the deterministic map that returns the missingness pattern of its input, so that $M^i=\mathsf m(X^i)$ almost surely; we use $M$ when emphasizing the random variable and $\mathsf m(\cdot)$ when emphasizing it as a function of the (masked) input.

Let $\obs(M)$ \footnote{ For example, conditionally on $M=[0,1,0]$,  $X=[X_1, \NA, X_3]$ thus  $\obs(M)=\{1,3\}$, $X_{\obs(M)}=[X_1,X_3]$.}
denote the set of indices of observed covariates and $\mis(M)$ that of the missing ones.
Given a new test point $X^{n+1}$, we seek an adaptive prediction interval $\hat{C}_\alpha$ such that 
\begin{equation}
	\label{eq:cond_cov}
\mbbP (Y^{n+1} \in \hat{C}_\alpha(X^{n+1}) \mid \mathsf{Cond}) \geq 1-\alpha-
\varepsilon(\mathsf{Cond})
\end{equation}
with $\varepsilon(\cdot)$, an error that can be controlled, and a conditioning event $\mathsf{Cond}$ defined later. Note that $X$  includes $\NA$s
and the observed data points are denoted by $Z^i = (X^i, Y^i)$.

We split the data into a \emph{training} subset with index $\mathcal I_{1}$ and a \emph{calibration} subset $\mathcal I_{2}$. 
Conformal wrappers require the choice of a \emph{nonconformity score} $S: \mathcal{X} \times \mY \rightarrow \mbbR$. Given an input \( X  \in \mathcal{X} \), $S$  quantifies the deviation between the observed label and the base prediction. 
Based on the calibration data $Z^{\mI_2}$ and the test point $Z^{n+1}$, we consider a \textit{score threshold function} $q: \mathcal{X} \rightarrow \mbbR$, which defines $\hat{C}(X)=\{y \in \mY: S(X, y) \leq q(X)\}.$
The standard split conformal prediction (SCP) procedure constructs an estimate of a constant $q(X)$ by the finite-sample-corrected $1-\alpha$ empirical quantile of scores $\{S(X^i, Y^i)\}_{i \in \mI_2}$ to achieve marginal coverage for $\hat{C}_\alpha$. Going beyond marginal coverage, however, this procedure may fail when there is missing data, as the conditional distribution of \( S \mid X \) depends on the mask \( M \). This motivates our extension toward mask-conditional in Section~\ref{sec:methodology}.

\section{Methodology}
\label{sec:methodology}
We assume $\mY=\mbbR$, focusing on the regression. 
Nevertheless, the theoretical results extend to classification as well, 
since nonconformity scores can be defined analogously for discrete spaces. We develop a new approach to identifying the score threshold function that is adaptive to masks.

As in standard split conformal analysis, the base model and representation are learned on the training split $\mathcal I_1$ and then treated as fixed during calibration and testing. Therefore, all theoretical statements and assumptions below are conditional on the training split.

Let $f_{\text{tree}}\colon\mathcal{X}\to\mathbb R$ be a quantile–regression forest or gradient–boosted tree ensemble trained on $Z^{\mathcal I_1}$ with $n_{\text{tree}}$ trees,
\(
f_{\text{tree}}(x)=\sum_{t=1}^{n_{\text{tree}}}\lambda_t\,T^{(t)}(x),
\)
where $\lambda_t\in\mathbb R$ are weights of each tree and  
$T^{(t)}\colon\mathcal{X}\to\mathbb R$ maps the masked input $x$ to the corresponding leaf value of the $t$-th tree.
The ensemble induces a deterministic \emph{tree embedding}
$
\whT(x)=\bigl[T^{(1)}(x),\ldots,T^{(n_{\text{tree}})}(x)\bigr]\in\mathbb R^{n_{\text{tree}}}.
$
The missing pattern of $X$ is denoted by $\mathsf{m}(x)$, and it is a binary vector in $\{0,1\}^d$ with $1$ at the coordinates where $x_j = \mathrm{NA}$. We augment the tree embedding with the raw mask and an intercept to define the linear function space:
\begin{equation}
	\mathcal{G} = \Big\{
	g_\beta(X) = \whT(X)\beta_1 + \mathsf{m}(X) \beta_2 + \beta_3 :  
	 \beta=[\beta_1, \beta_2, \beta_3] \in \mathbb{R}^{n_{\tree}} \times \mbbR^d \times \mbbR
	\Big\}
\end{equation}
and consider the following assumption.
\begin{assumption}
	\label[assumption]{assump:linearind}
	The set of scalar basis functions $\Bigl\{\whT(\cdot),\; \mathsf{m}(\cdot),\; 1\Bigr\}$
	is linearly independent in $L^2(\mbbP_X)$.
\end{assumption}

Assumption~\ref{assump:linearind} is a non-degeneracy condition for the basis in $L^2(\mathbb{P})$ which only avoids the almost sure linear relation such as $\sum _i a _i T^{(i)}(X) + \sum _j b _j \mathsf{m} _j(X) + c = 0$ and does not violate the general missingness mechanism.
In particular, the training procedure of tree ensembles prevents two distinct leaves from being exactly identical (as this would yield no further information gain), 
and each leaf results from multiple splits of $X$, making it highly unlikely to coincide with a single indicator $\mathsf{m}_j$. Under \cref{assump:linearind}, we introduce a scalar product on $\mathcal G$ by
$\langle g_\beta, g_{\beta'} \rangle := \langle \beta, \beta' \rangle$,
and equip $\mathcal G$ with the corresponding norm $\|g_\beta\| :=\|\beta\|_2$. In practice, both the tree embedding $\widehat T(x)$ and the mask features $\mathsf{m}(x)$ are standardized before fitting, making their scales comparable. 

\paragraph{Fitting via regularized pinball loss.}
Let $\{(X^i,S^i)\}_{i \in \mI_2}$ be the calibration points, where
$S^i=S(X^i,Y^i)$ is any nonconformity score.  
For any new observation $X^{n+1}$ and any candidate score $s\in\mathbb R$, we define the regularized empirical risk
\begin{align}
	\label{eq:regularised loss}
	\hat{L}^s_\lambda(g_\beta)\;=\;\frac{1}{|\mI_2|+1} \!\!\sum_{i \in \mI_2} \ell_{\alpha}\!\bigl(
	g_\beta(X^i), S^{i}\bigr) \;+\;\frac{1}{|\mI_2|+1}\, \ell_{\alpha}\!\bigl(g_\beta(X^{n+1}), s\bigr)
	\;+\;
	\lambda\, \mR(g_\beta) ,
\end{align}

where the $\alpha$-pinball loss is $\ell_{\alpha}(\theta,s)
=
\frac{1}{2}|s-\theta|
+
\left(\frac{1}{2}-\alpha\right)(s-\theta)$
and the penalty is defined as $\mathcal R(g_\beta) = \lVert J \beta \rVert_2^2,$
with $J \in \mathbb{R}^{r \times (n_{\tree}+d+1)}$ a fixed design matrix, where $r$ is the row dimension of $J$.  Then we solve the following optimization problem
\begin{align}
	{g}^s_{\hat{\beta}}
	&\;=\; \underset{g_\beta \in \mG}{\operatorname{argmin}} \; \hat{L}_{\lambda}^s(g_\beta)\,.
	\label{eq:minimizer}
\end{align}

The matrix $J$ allows us to flexibly specify which directions in the parameter space are penalized:
For example, $J=I$ yields the standard ridge penalty, while a diagonal or structured $J$ can impose heterogeneous shrinkage. In our analysis, different choices of $J$ correspond to different targets.

Following \citet{gibbs2025conformal}, we define the conformal prediction set for the test point as
\begin{equation}
	\label{eq:final-set}
	\hat C_{\alpha}(X^{n+1})
	\;:=\;
	\bigl\{y\in\mathcal Y:\;
	S(X^{n+1},y)\le g^{S(X^{n+1},y)}_{\hat\beta}(X^{n+1})\bigr\}.
\end{equation}

Of course, the dependence of $g^{s}_{\hat\beta}$ on the candidate score $s$ makes the prediction set implicit, but we will show below that the construction \eqref{eq:minimizer} satisfies strong duality and can be solved efficiently by the dual method of \citet{gibbs2025conformal}. In the remainder of this section, we derive guarantees corresponding to different choices of the conditioning event $\mathsf{Cond}$ in \cref{eq:cond_cov}.

\subsection{$\mathsf{Cond}: \mathsf{m}_j(X) = 1,\; $ $\textnormal{ for any fixed } j \in \{1, \dots, d\}$}

Missingness patterns directly affect the conditional distribution of the response, as the absence of certain coordinates reduces the information available for prediction and typically increases the variability of $Y \mid X$. 
We first target \emph{coordinate-wise} missingness-conditional coverage, meaning coverage conditional on the event that one specified feature is missing. This first target is useful when certain covariates are clinically important, while avoiding the intractability of enforcing exact coverage over exponentially many full masks.
To derive our exact guarantees, we require \cref{assump:symmetric}, which is standard and satisfied by many machine learning algorithms (e.g., linear regression or forest models). Without loss of generality, we assume in \cref{assump:mpropa} that all masks occur with probability non-zero in \cref{assump:mpropa}; otherwise, adaptive guarantees would be vacuous. 
\begin{assumption}
	\label[assumption]{assump:symmetric}
	The optimization procedure that yields $g_{\hat\beta}$,
	including the selection of a minimizer
	is symmetric with respect to the calibration points
	together with the candidate test point.
\end{assumption}
\begin{assumption}
	\label[assumption]{assump:mpropa}
	For all $m \in \{0,1\}^d$, $\mathbb P(M=m) > 0$.
\end{assumption}

\begin{theorem}
	\label{thm:coord_mask_cond}
	Let $\{(X^i, Y^i)\}_{i=1}^{n+1}$ be i.i.d. samples, with missingness pattern  
	$M^i=\mathsf{m}(X^i)$. Let $S(X, Y)$ be a nonconformity score, and let ${g}^s_{\hat{\beta}} \in \mathcal{G}$ be the fitted quantile function from \Cref{eq:minimizer}, with $J=J_1 := \mathrm{diag}(1,\ldots,1,0,\ldots,0)$,
	where the first $n_{\text{tree}}$ entries equal 1 and the last $d+1$ entries equal 0. Under \cref{assump:symmetric,assump:mpropa}:
	\(\forall  j \in \{1, \dots, d\}  \), the conformal prediction interval $\hat{C}_\alpha(X)$ defined in \Cref{eq:final-set} achieves the coordinate-wise conditional coverage:
	\begin{equation}
		\mathbb{P}\left( Y^{n+1} \in \hat{C}_\alpha(X^{n+1}) \mid \mathsf{m}(X^{n+1})_j = 1 \right)
		\ge 1 - \alpha.
	\end{equation}
\end{theorem}

The proof is in \cref{proof:thm1}. \Cref{thm:coord_mask_cond} shows that our procedure adapts to the missingness mechanism at the level of individual coordinates: whenever feature $j$ is missing, the resulting interval maintains the target coverage.

\subsection{$\mathsf{Cond}: M=m, \; \forall m \in \mathcal{M}_{\mathrm{cal}}$}
\label{subsec:obsmask}

In high-dimensional settings with missing values, the total number of possible missingness patterns is $2^d$, and thus requiring exact Mask-Conditional Coverage (MCV) for \emph{every} $m \in \{0,1\}^d$ is difficult, as many patterns may simply never appear in the finite sample. A practical intermediate goal is to ensure validity for the specific patterns observed in the available data.

Let $\mathcal{M}_{\mathrm{cal}}:= \{ \mathsf{m}(X^i) : i \in \mathcal{I}_2 \}$ denote the set of unique missingness patterns in the calibration set. To guarantee coverage conditioned on those masks, we explicitly augment our hypothesis space $\mathcal{G}$ with indicator functions for each observed pattern. Specifically, we use the representation $g'_\beta \in \mathcal{G}_{\mathrm{cal}}$: 
\begin{align*}
	g'_\beta(X) &= \whT(X)\beta_1
	+ \sum_{m \in \mathcal{M}_{\mathrm{cal}}} \beta_m \ind\{\mathsf{m}(X) = m\} + \beta_0.
\end{align*}
To achieve the coverage, we exclude the coefficients $\{\beta_m\}_{m \in \mathcal{M}_{\mathrm{cal}}}$ from the regularization penalty, by setting the corresponding diagonal value in  $J$ to zero. 
For the validity argument, since we use calibration data to form function space, it is thus not a direct application of the first-order optimality as in Theorem~\ref{thm:coord_mask_cond}. We use the permutation-invariant auxiliary dictionary indexed by $\mathcal M_{\mathrm{aug}}=\mathcal M_{\mathrm{cal}}\cup\{M^{n+1}\}$, which agrees with the original dictionary whenever the test mask has been observed in calibration.
\begin{proposition}
	\label{prop:calib_mask_cond}
	Under the same assumptions as Theorem~\ref{thm:coord_mask_cond}, with the symmetry condition applied to the auxiliary augmented procedure and $|\mathcal I_2|\ge1$, suppose we solve the quantile regression criterion~\eqref{eq:minimizer} with the hypothesis space $\mathcal{G}_{\mathrm{cal}}$ instead of $\mathcal{G}$ and penalty $J=J_2 := \mathrm{diag}(1,\ldots,1,0,\ldots,0)$, where the first $n_{\text{tree}}$ entries equal 1 and the rest are 0. Then for every fixed mask $m\in\{0,1\}^d$, under the conditional law given $m\in\mathcal M_{\mathrm{cal}}$, the prediction set satisfies:
	\begin{equation}
		\mathbb{P}\left( Y^{n+1} \in \hat{C}_\alpha(X^{n+1}) \mid M^{n+1} = m \right) \ge 1 - \alpha.
	\end{equation}
\end{proposition}
The proof is in Appendix~\ref{proof:observed-mask}.  This result covers masks supported by the calibration sample.  For unseen patterns ($m \notin \mathcal{M}_{\mathrm{cal}}$), validity must rely on how well the representation generalizes across masks, which is the focus of the next subsection.

\subsection{$\mathsf{Cond}:M=m, \; \forall m \in \{0,1\}^d$}
Generally, the previous coordinate coverage $M_j = m_j$ does not imply MCV. 
To study general mask-conditional results, we return to the augmented class $\mathcal G$ and analyze the tree-induced feature space. This yields an approximate guarantee for each fixed mask, with an error term reflecting how well the mask indicator can be approximated by learned representations. For theoretical analysis, we consider the $K$-bounded space
$
	\mG_K=\{g_\beta \in \mG: \|\beta\|_2 \leq K\}.
$

For a fixed mask $m\in\{0,1\}^d$, let $\pi_m:=\mathbb P(M=m)>0$, $\rho_m(x):=\ind\{\mathsf m(x)=m\}$, and $r_m(x):=\rho_m(x)/\pi_m$ be its normalized indicator. Let $r_K\in\mathcal G_K$ be the $L^2(\mathbb P_X)$-projection of $r_m$ on $\mathcal G_K$, and define $\Delta_{K,m}:=\|r_m-r_K\|_{L^2(\mathbb P_X)}$. This representation error measures how well the shared tree-plus-mask class approximates the normalized mask indicator.

We assume the following condition on the score distribution around the threshold class, together with the full-rank penalty in \cref{assump:strong-convex}, bounded scores in \cref{assump:bounded-scores}, and a well-conditioned augmented feature Gram matrix in \cref{assump:gram}.
\begin{assumption}
	\label[assumption]{assump:anti-concentration}
	There exist constants $C_{0}>0$ and $t_0>0$ such that, for every
	$\beta \in B_K := \{\beta:\|\beta\|_2\le K\}$ and every $t \in (0,t_0]$,
	\(
	\mbbP\!\left(
	|S(X,Y)-g_{\beta}(X)|\le t
	\right)
	\le C_{0}t .
	\)
\end{assumption}

The full-rank penalty gives strong convexity, while bounded scores and features control the loss. Although the pinball loss is nonsmooth, the above assumption makes the population risk smooth and controls changes in coverage under small threshold perturbations. We introduce the notation needed for the statement of the following theorem. Write $\psi(x)=[\widehat T(x),\mathsf m(x),1]$. The constants $B_\psi$ and $B_\ell$ bound $\|\psi(X)\|_2$ and the pinball loss over $\mathcal G_K$, respectively; $\lambda_J$ and $\lambda_{J\mathrm{max}}$ are the smallest and largest eigenvalues of $J^\top J$. Let $L(g_\beta):=\mathbb E[\ell_\alpha(g_\beta(X),S)]+\lambda\|J\beta\|_2^2$ denote the population risk. Set $u:=|\mathcal I_2|+1$. See Appendix~\ref{sec:generror} for the auxiliary assumptions.

\begin{theorem}
	\label{thm:mask-cond-generalJ-revised}
	Suppose Assumptions~\ref{assump:linearind}, \ref{assump:symmetric}, \ref{assump:mpropa},\ref{assump:anti-concentration}, \ref{assump:strong-convex}, \ref{assump:bounded-scores}, and \ref{assump:gram} hold. Assume $\alpha\le 0.5$ and the constrained population minimizer $\beta^\star\in\arg\min_{\beta\in B_K}L(g_\beta)$ lies in the interior of $B_K$ (which can typically be ensured by taking $K$ or $\lambda$ sufficiently large), with $\kappa=K-\|\beta^\star\|_2>0$.
	Use the candidate-augmented optimizer over $\mathcal G_K$ in~\eqref{eq:final-set}.
	Then there exist constants $C_1,C_2>0$ depending only on $\alpha,B_\psi,B_\ell,K,\kappa,\lambda_J,\lambda_{J\mathrm{max}},C_0,t_0$, such that, for any $\delta\in(0,1)$,
	with probability at least $1-\delta$ over the draw of the calibration sample $\mathcal C$,  we have under the conditional law given $\mathcal C$:
	\begin{align}
		\label{eq:mcovsimplified}
		\mathbb{P}\!\left(Y\in\widehat C_\alpha(X)\,\middle|\,M=m\right)
		\ge
		1-\alpha
		-
		C_{1}\frac{(1+\sqrt{\lambda})(\log(2/\delta))^{1/4}}
		{\sqrt{\lambda}\,u^{1/4}}
		-
		C_{2}\sqrt{\lambda}
		-
		(1-\alpha)\Delta_{K,m}.
	\end{align}
	
\end{theorem}

\paragraph{Proof sketch.}  We first fit the regularized objective using only the calibration $\mathcal C$. Stability bounds its population excess risk, population smoothness then controls the gradient in the direction $r_K$. The population-gradient bounds the projected coverage moment, with a regularization error. Add-one stability and \cref{assump:anti-concentration} transfer this moment to the candidate-augmented fit. Finally, Cauchy--Schwarz bounds the contribution of $r_m-r_K$ by $(1-\alpha)\Delta_{K,m}$. Theorem~\ref{thm:mask-cond-generalJ_app} in Appendix~\ref{sec:generror} gives the complete bound and proof. 
\begin{remark}[Estimation of $\Delta_{K,m}$]
	\label{remark:estimation}
	$\Delta_{K,m}$ is the $L^2(\mathbb P)$ projection error, i.e., $\Delta_{K,m}=\|r_m-r_K^\star\|_{L^2(\mathbb P)}$ where $r_K^\star$ is the $L^2$ projection of $r_m$ onto $\mathcal G_K$. With finite samples, one can estimate it by fitting $r_m(x)$ with (regularized) least squares over $\mathcal G_K$  and then use the empirical residual as a diagnostic. 
\end{remark}

\subsection{Efficient Computation via Dual Formulation}
\label{subsec:dual}
A difficulty of solving \cref{eq:final-set} is the non-explicit form of the prediction set, which is computationally prohibitive. It is impossible to solve for any $y$(or score $S(X,y)$) a new regularized minimization problem. Following \citet{gibbs2025conformal}, we will prove that our new tree-embedding-based pinball loss satisfies the strong duality condition and can be solved efficiently.

\begin{proposition}
	\label{prop:strong-duality}
	Let $u=|\mathcal I_2|+1$. Assume $J^\top J\succeq0$ and $\lambda>0$.
	For $\eta$ indexed by $\mathcal I_2\cup\{n+1\}$, define
	\(
	\mathcal R_{\mathcal I_2,n+1}^\star(\eta)
	:=
	-
	\inf_{\beta}
	\left\{
	u\lambda\|J\beta\|_2^2
	-
	\sum_{i\in\mathcal I_2}\eta_i g_\beta(X^i)
	-
	\eta_{n+1}g_\beta(X^{n+1})
	\right\}.
	\)
	Then the slack-variable form of \eqref{eq:minimizer} is a finite-dimensional
	convex program satisfying Slater's condition, and thus strong duality holds.
	Its dual is
	\(
	\max_{\eta}
	\left\{
	\sum_{i\in\mathcal I_2}\eta_i S^i
	+
	\eta_{n+1}s
	-
	\mathcal R_{\mathcal I_2,n+1}^\star(\eta)
	\right\}\)
	s.t. 
	\(
	-\alpha\le\eta_i\le1-\alpha,
	\text{ for } i\in\mathcal I_2\cup\{n+1\}.
	\)
\end{proposition}

See \cref{proof:strong duality} for proof. For each candidate score $s$, let
$\eta^\star(s)$ be an optimizer of the dual problem in
Proposition~\ref{prop:strong-duality}, and write $\eta^\star_{n+1}(s)$ for the
coordinate associated with the test point. By the KKT conditions, testing the
primal inclusion condition $s \le g^s_{\hat\beta}(X^{n+1})$ in
\cref{eq:final-set} is equivalent to
testing whether $\eta^\star_{n+1}(s)<1-\alpha$. Moreover, by the monotonicity
result of Theorem~4 in \citet{gibbs2025conformal}, $\eta^\star_{n+1}(s)$ is
non-decreasing in $s$. Therefore, for the test point $X^{n+1}$, the prediction
set can be computed as
\(
\hat C_\alpha(X^{n+1})
=
\{y:S(X^{n+1},y)\le s^\star\},
\)
where
\(
s^\star=\sup\{s:\eta^\star_{n+1}(s)<1-\alpha\}.
\)
For this, standard solvers (QP, bisection, secant) converge rapidly.

\section{Experiments}
We use the conformalized quantile regression (CQR) score $S(x,y)=\max\{\hat{\mu}_{\mathrm l}(x)-y,\;y-\hat{\mu}_{\mathrm u}(x)\}$, where the lower and upper quantile estimators are trained directly on masked covariates.
Both $\hat{\mu}_{\mathrm l}$ and $\hat{\mu}_{\mathrm u}$ are HistGBDT quantile estimators \citep{ke2017lightgbm} learned on the training split, and no imputation is used before fitting these base models.
The target level is fixed at $1-\alpha=90\%$. All base-model fitting, representation learning, preprocessing, and tuning use proper training data only. In each replication, the training pool is split equally into base-fit and selection halves. The latter supplies validation and five-fold cross-validation for $\lambda$. The calibration split is then used only for final conformal fitting. 

The experiments are organized around the three validity targets in Section~\ref{sec:methodology}.
{First, coordinate-wise and observed-mask experiments directly evaluate subgroup coverage for the exact finite-sample targets.}
{Second, global comparisons with (MDA-)Exact, Nested \citep{zaffran2023conformal}, Weighted CP, and acceptance-rejection corrected (ARC) CP \citep{fan2025arxiv} use aggregate mask-wise shortfall risk and interval length, because no compared method can provide exact validity for every possible mask under arbitrary mechanisms.}
Third, we include ablations to identify the roles of tree representations and rare masks. 
For fairness, all methods use the same train/calibration/test splits and the same HistGBDT base quantile model $(\hat{\mu}_l,\hat{\mu}_u)$; HistGBDT hyperparameters are tuned by Optuna \citep{akiba2019optuna} with 100 trials.

\subsection{Synthetic dataset}
We generate covariates \( X \in \mathbb{R}^d \) from a multivariate Gaussian distribution \( X \sim \mathcal{N}(\nu, \Sigma) \), where \( \nu = \mathbf{1}_d \) and the covariance matrix is defined as
$
\Sigma = \rho \cdot \mathbf{1}_d \mathbf{1}_d^\top + (1 - \rho) I_d,
$
with \( \rho = 0.8 \). The response variable \( Y \) follows a linear model:
$
Y = \beta^\top X + \varepsilon, \quad \varepsilon \sim \mathcal{N}(0, 1)$, $\varepsilon \perp\!\!\!\perp (X, M).
$
The regression coefficient \( \beta \in \mathbb{R}^{d} \) is fixed as $
\beta_{d=3} = (1,\; 2,\; -1)^\top,
$ and $\beta_{d=6} = (1,\; 2,\; -1, 3,\; -0.5,\; -1)^\top,$ for 2 different dimension \( d \in \{3,6\} \) of dataset. 
Each dataset is split into 1000 samples for training the base model and 500 samples for calibration. 

\paragraph{Missingness Mechanisms.} We consider three common missingness settings: 
(1) \textbf{MCAR \& MAR:}\citep{zaffran2023conformal,mayer2019r} We adopt standard MCAR and MAR settings as baselines. In MCAR, the missingness probability is constant across all samples. In MAR, the missingness of a variable depends on the values of fully observed covariates via a logistic link function. (2)  \textbf{Pattern Mixture (PMIX)} (see e.g. \citet{ayme2022near} or Model 4 in \citet{little2008selection}): We introduce this Missing Not at Random (MNAR) setting based on a Pattern Mixture Model, which models the dependence of the covariate distribution on the missingness mask $M$. Specifically, for a given mask $m \in \{0,1\}^d$, we sample covariates from a conditional Gaussian distribution:
$X \mid M=m \;\sim\; \mathcal{N}(\mu(m), \Sigma(m))$,
where the mean $\mu(m) = \mu_0 + B m$ shifts linearly with the mask vector, and the covariance $\Sigma(m) = (s_0 + s_1 \|m\|_1) \Sigma_0$ scales according to the number of missing features $\|m\|_1$. This induces  \emph{mask-conditional heterogeneity}. This setting provides a rigorous stress test for our method's ability to deliver valid mask-conditional coverage.

\paragraph{Evaluation Metrics.} Each experiment is repeated over 10 random seeds to estimate empirical coverage and average interval length with 95\% confidence intervals across seeds. To validate the coordinate-wise conditional guarantee and the observed mask conditional guarantee, it is natural to report pattern-wise coverage. To compare all mask-conditional coverage, we use the two metrics: 
(1) {Coverage Shortfall Risk:} defined as 
$\frac{1}{\#\mathcal{M}_{\text{test}}}\sum_{m\in \mathcal{M}_{\text{test}}}[\max((1-\alpha) - \hat{\mathbb{P}}(Y \in \hat{C}_\alpha(X)|M=m), 0)],$ 
where $\mathcal{M}_{\mathrm{test}}$ is the set of all masks observed on the test set. Unlike raw coverage, this metric penalizes violations of the target coverage, which we view as a critical indicator of safety in practice.
(2) {Interval Length:} The average length of the prediction sets, reflecting informativeness.

\paragraph{Coordinate-wise Conditional Coverage}

\begin{figure}[t]
	\centering
	\begin{minipage}[t]{0.45\linewidth}
		\centering
		\includegraphics[width=\linewidth]{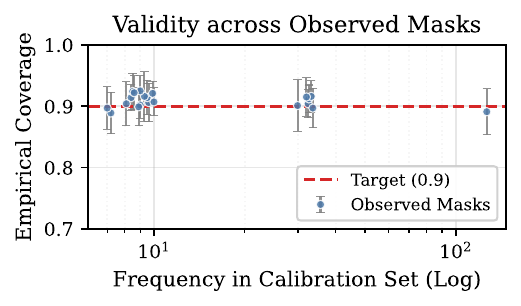}
		\caption{\textbf{Validity across observed masks.} The scatter plot shows the empirical coverage on 100 stratified test points for every mask pattern that exists in the calibration set.}
		\label{fig:coverage_vs_freq}
	\end{minipage}
	\hfill
	\begin{minipage}[t]{0.45\linewidth}
		\centering
		\includegraphics[width=\linewidth]{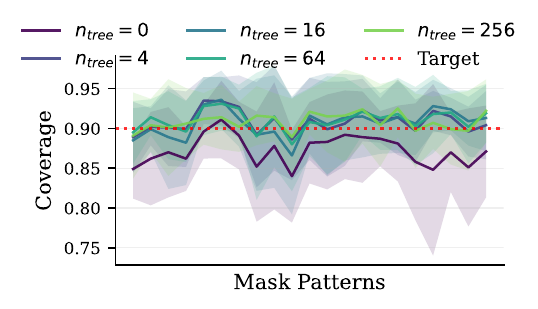}
		\caption{\textbf{Sensitivity to $\mathbf{n_{\textbf{tree}}}$.} Coverage across mask in synthetic MAR setting ($d=6$). \textit{A clear miscoverage when {no tree}, showing {the importance of tree embeddings.}}}
		\label{fig:dim6_ntree}
	\end{minipage}
\end{figure}

We empirically validate the coordinate-wise conditional coverage guarantee derived in Theorem~\ref{thm:coord_mask_cond}, which ensures that valid coverage is maintained conditionally on specific missing features, i.e., $\mathbb{P}(Y \in \hat{C}_{\alpha}(X) \mid M_j = 1) \ge 1 - \alpha$.
Table~\ref{tab:coord_coverage} reports results on the $d=6$ synthetic dataset at nominal level $90\%$. Under MCAR and PMIX, missingness can occur across all dimensions, 
and under MAR, only the first three features can be missing by construction, so the events $M_j=1$ for $j\in\{4,5,6\}$ do not occur.  We observe that the empirical coverage remains close to the nominal level without being overly conservative.  
\begin{table}[t]
	\centering
	\caption{Coordinate-wise coverage estimates with $d=6$ and nominal level $90\%$. Entries are mean percentage coverage $\pm$ 95\% CI half-width over 10 seeds.}
	\label{tab:coord_coverage}
	\begin{tabular}{lcccccc}
		\toprule
		\textbf{Mechanism}
		& $M_1=1$ & $M_2=1$ & $M_3=1$ & $M_4=1$ & $M_5=1$ & $M_6=1$ \\
		\midrule
		\textbf{MCAR} & $92.3\pm2.1$ & $92.2\pm1.6$ & $91.7\pm2.0$ & $92.4\pm1.1$ & $91.5\pm2.2$ & $92.1\pm2.2$ \\
		\textbf{MAR}  & $92.8\pm1.1$ & $92.7\pm1.9$ & $92.6\pm1.8$ & N/A & N/A & N/A \\
		\textbf{PMIX} & $91.7\pm1.9$ & $92.1\pm2.3$ & $90.8\pm2.0$ & $93.9\pm1.1$ & $92.4\pm2.2$ & $92.9\pm0.9$ \\
		\bottomrule
	\end{tabular}
\end{table}
\paragraph{Exact Coverage on Observed Patterns.}
Beyond coordinate-wise validity, we further verify Proposition~\ref{prop:calib_mask_cond}, which guarantees exact coverage for \emph{any} specific missingness pattern $m$ present in the calibration set $\mathcal{M}_{\mathrm{cal}}$.
To rigorously validate this, we use a stratified test set (100 independent samples per mask) to compute the empirical coverage for \textit{every unique mask} observed in the calibration data.
Figure~\ref{fig:coverage_vs_freq} shows these results.
For masks that appear in the calibration set, the coverage is close to the nominal $1-\alpha$ level.
Crucially, even for rare masks (e.g., those appearing fewer than 10 times), coverage remains centered around the target.
This empirically confirms that our method successfully activates the unpenalized mask indicators, ensuring group-conditional validity across all observed regimes.

\begin{figure}[t]
	\centering
	\begin{subfigure}[t]{0.49\linewidth}
		\centering
		\includegraphics[width=\linewidth]{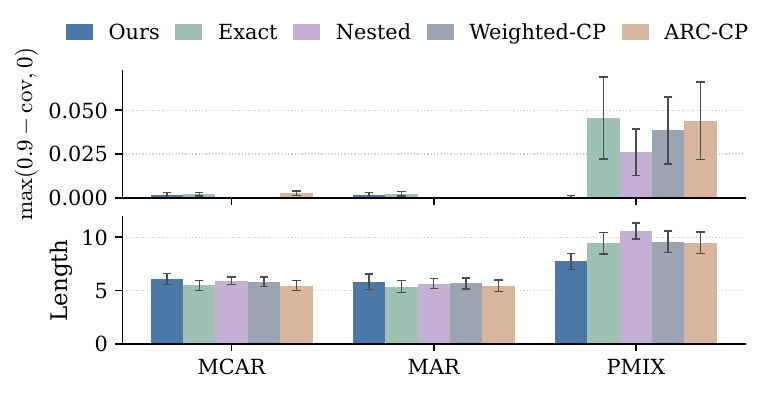}
		\caption{$d=3$, $p=0.2$.}
		\label{fig:dim3-shortfall}
	\end{subfigure}
	\hfill
	\begin{subfigure}[t]{0.49\linewidth}
		\centering
		\includegraphics[width=\linewidth]{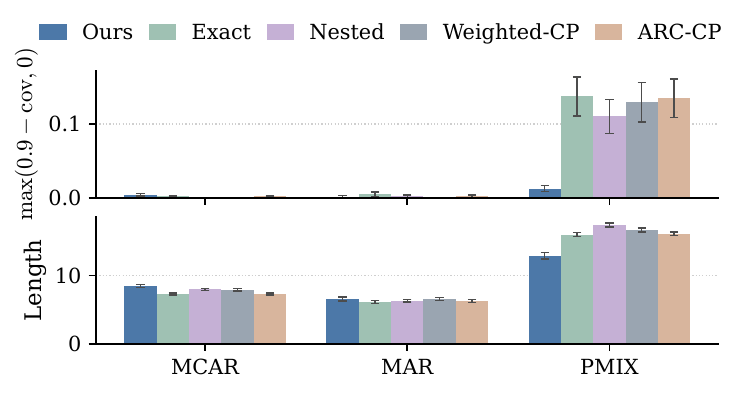}
		\caption{$d=6$, $p=0.2$.}
		\label{fig:dim6-shortfall}
	\end{subfigure}
	
	\caption{
			Empirical shortfall risk and average interval length across missingness mechanisms.
		Our method is competitive with the baselines on MCAR/MAR and gives a stronger coverage--length trade-off on the mask-heterogeneous PMIX setting.
		Here $n_{\mathrm{tree}}=256$.
	}
	\label{fig:shortfall}
\end{figure}

\paragraph{Global Validity and Efficiency.}
Figure~\ref{fig:dim3-shortfall} presents the global performance for the 3-dimensional dataset. Our method achieves comparable shortfall risk and interval length to baselines in the MCAR and MAR settings. 
Under PMIX, where the conditional distribution of $X$ changes substantially with the mask, our method maintains low shortfall risk while producing shorter intervals than the more conservative baselines. Figure~\ref{fig:dim6-shortfall} shows the same qualitative pattern in $d=6$.
In practice, we do not impose an explicit bound constraint $\|\beta\|_2 \le K$. For full-rank ridge, the deterministic bound $\|\widehat\beta\|_2\le C_\ell B_\psi/(2\lambda\lambda_J)$ supplies a sufficient radius. This is also detailed in Appendix~\ref{sec:rate-explicit}.
\paragraph{Importance of Tree Embeddings.}
To assess the benefit of the non-linear feature representations learned by the tree ensemble, we evaluate performance as a function of the number of trees \( n_{\text{tree}} \) in the base quantile regressor. Figure~\ref{fig:dim6_ntree} presents the results for the MAR setting ($d=6$).
We observe a clear trend: with no trees, coverage drops significantly below the target. This finding shows the limitations of the mask-only linear representation and highlights the importance of \emph{tree embeddings} for capturing mask-conditional uncertainty in this setting.
Notably, performance stabilizes rapidly once $n_{\text{tree}} \ge 10$. 
This suggests that our method is robust to hyperparameter selection: by relying on cross-validation to select $\lambda$, users can employ a sufficiently large forest without extensive tuning, reliably achieving both valid coverage and informative intervals.

\subsection{High (100)-dimensional  Dataset}

From this section, we focus on arbitrary mask coverage of \cref{thm:mask-cond-generalJ-revised}  and consider the function space $\mathcal{G}$ with $n_\tree=256$ for all following experiments. To assess our scalability, we extend the synthetic regression task to a high-dimensional setting with $d=100$.  
The space of all missingness patterns ($2^{100}$) is large, making it computationally intractable to evaluate all masks. Therefore, we randomly sample a fixed set of $N_{\text{mask}}=100$ distinct missingness masks. This experiment uses 5000 training samples and 1000 calibration samples.
For each sampled mask, we generate 100 test points to form a representative test set.
To ensure a fair comparison, this set of sampled masks is \textbf{fixed} across all evaluated methods.

This experiment tests \textbf{high-dimensional and rare-mask generalization}: it evaluates whether tree-based embeddings can effectively learn the structure of missingness representations, and almost all the evaluated masks are unseen masks in training and calibration. Due to space, the results are shown in the Appendix in Figure \ref{fig:dim100-shortfall}. Nested is not listed because it cannot finish in a reasonable time. Our method is the only evaluated method on PMIX that achieves near-zero empirical shortfall risk while keeping the prediction intervals informative.

\subsection{Semi-Synthetic Real World Dataset}
We consider a popular benchmark dataset, Concrete Compressive Strength, in related work \citep{zaffran2023conformal, fan2025arxiv}. This dataset contains 8 continuous features and captures the highly nonlinear relationship between concrete ingredients, age, and resulting structural strength. We use 630 training samples, 200 calibration samples, and a pattern-stratified test set with 100 masks and 100 samples per mask. We allow all features to be missing with 20\% probability, and we consider MCAR, MAR, and self-masked MNAR 
\citep{zaffran2023conformal,mayer2019r} for the missingness mechanism. The self-masked MNAR is similar to MAR, but the probability of missing each covariate depends on its own original value.  
Figure \ref{fig:concrete-shortfall} shows the comparison of all methods. Our method remains the only one to achieve near-zero risk across all missingness, especially in a self-masked MNAR setting, where competing methods suffer from miscoverage. Although our intervals are wider, they remain informative and reflect the expected trade-off between validity and interval length.

\subsection{Real World Dataset (MIMIC-IV LOS)}
We evaluate our method on a real-world regression task constructed from the MIMIC-IV dataset \citep{PhysioNet-mimiciv-3.1, johnson2023mimic} and follow its benchmark pipeline \citep{Harutyunyan2019}.
The underlying task is \emph{length-of-stay (LOS) forecasting} in the ICU.
The covariates are computed by summarizing statistics over sub-windows (min/max/mean/std/skew/count); this yields a high-dimensional tabular representation with missing values due to irregular measurements. A simple feature selection step is applied to the training split using permutation feature importance with a HistGradientBoostingRegressor model.
This results in a $d=300$-dimensional regression dataset with missingness.

\begin{figure}[t]
	\centering
	
	\begin{minipage}[c]{0.53\linewidth}
		\centering
		\includegraphics[width=\linewidth]{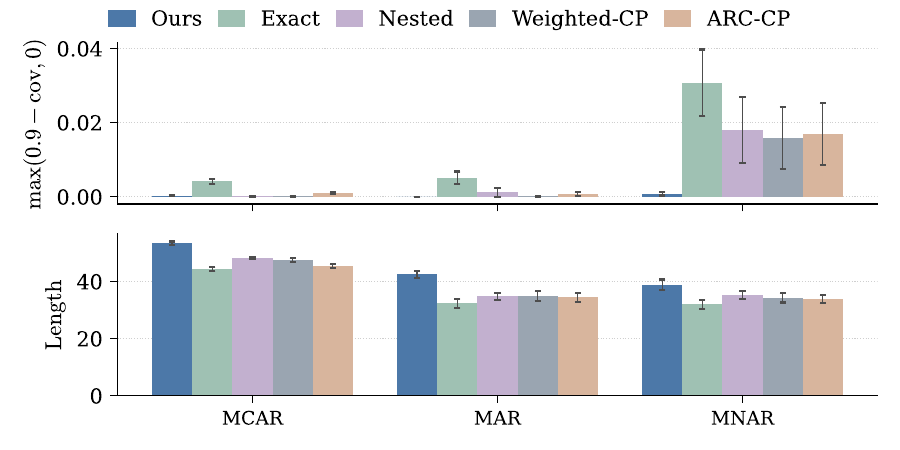}
		\caption{
			Empirical shortfall risk and average interval length on the Concrete dataset. Our method achieves the lowest shortfall risk among the evaluated methods.
		}
		\label{fig:concrete-shortfall}
	\end{minipage}
	\hfill
	\begin{minipage}[c]{0.44\linewidth}
		\centering
		\includegraphics[width=\linewidth]{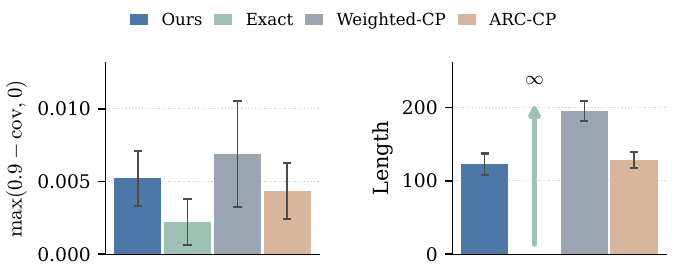}
		\includegraphics[width=.95\linewidth]{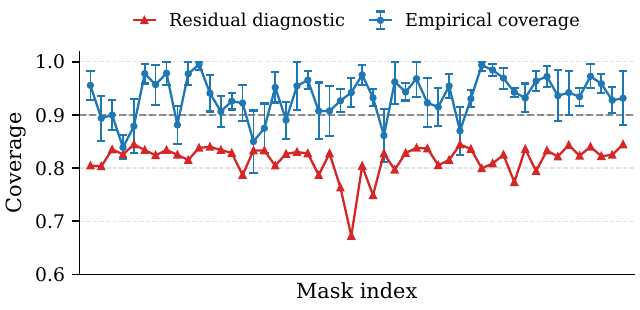}
		\captionof{figure}{
		MIMIC-IV LOS results. Top: coverage shortfall risk and average interval length. Bottom: mask-wise empirical coverage versus the residual diagnostic defined in the text.
		}
		\label{fig:mimic-combined}
	\end{minipage}
\end{figure}

For evaluation, we repeat each experiment across 10 random seeds, randomly drawing 3000 training samples and 1000 for calibration. Following the benchmark, we use the entire test set, which contains 2331 samples with 50 different missingness patterns.
All methods use the same data splits and the same base model, and we report empirical shortfall miscoverage risk and interval length at the nominal $1-\alpha=90\%$. Figure \ref{fig:mimic-combined} (top) compares our method to baselines on the MIMIC-IV LOS task. Due to computational overhead, the Nested-MDA baseline could not complete within a reasonable time, consistent with the high-dimensional synthetic experience. Exact-MDA produced extremely wide intervals (often diverging). 
Among the remaining baselines, our method achieves the smallest interval lengths while giving a slightly higher shortfall risk than the other two methods.
Notably, the average shortfall risk remains below $0.5\%$, which is modest relative to the reduction in interval length, highlighting a practical trade-off between conditional adaptivity and efficiency. The masks with coverage below 85\% exhibit low frequencies (1.58 times across 10 replications) in the calibration set. We observe from \cref{fig:mimic-combined} (bottom) that the miscoverage is mainly driven by a small number of mask patterns, while the majority of patterns maintain valid coverage.
We examine representation quality through the in-sample residual $\widehat\Delta_{\rho,m}$. The diagnostic is the per-mask minimum across seeds of $1-\alpha-c_\alpha\widehat\Delta_{\rho,m}$, clipped to $[0,1]$, with $c_\alpha=\max\{\alpha,1-\alpha\}$. This diagnostic isolates the representation component. See Remark~\ref{remark:estimation} for the statistical interpretation.

\subsection{Imputation-plus-mask baseline }
We replace only the tree embedding with MICE-imputed covariates concatenated with the raw mask and an intercept, retaining the same training-only tuning protocol. Table~\ref{tab:mice-matched} isolates this representation comparison. On PMIX, coverage is $0.8476$ versus $0.9145$ for tree-plus-mask; on MIMIC, MICE has larger shortfall at essentially the same interval length.
\begin{table}[t]
	\centering
	\caption{Matched MICE-plus-mask baseline (10-seed means; nominal coverage $0.9$). }
	\label{tab:mice-matched}
	\begin{tabular}{lcccc}
		\toprule
		Setting & Coverage & Shortfall & Interval length & Masks below $0.9$\\
		\midrule
		MCAR & 0.8613 & 0.0389 & 6.960 & $62/64$\\
		MAR & 0.8791 & 0.0209 & 5.345 & $8/8$\\
		PMIX & 0.8476 & 0.0552 & 10.371 & $51/64$\\
		MIMIC-IV & 0.8897 & 0.0493 & 115.040 & $27/50$\\
		\bottomrule
	\end{tabular}
\end{table}
\subsection{Running time analysis}
{We also report runtime in Appendix~\ref{sec:runtime}.}
{The exact dual method is slower than several baselines in low dimensions, whereas in high dimensions it remains feasible in settings where MDA-Nested does not finish within a reasonable time.}

\section{Conclusion and future work}
We focused on conformal prediction in the setting where covariates are missing. The missingness patterns create significant differences in distribution. We proposed a novel conformal framework that combines tree-based embeddings with quantile learning to specifically address missing values in any dataset. 
The theory gives finite-sample guarantees for coordinate-wise and observed-mask targets, and an approximate guarantee for all masks whose error depends on estimation, regularization, and representation quality.
Empirically, the method reduces mask-wise coverage-shortfall risk in challenging settings while maintaining informative intervals.

\paragraph{Limitations}
The guarantees in this paper are regime-specific. While our methods empirically perform well, the theoretical result is approximate and depends on the representation quality and the regularity conditions. Computationally, the exact dual path can be slower than several baselines. Extensions to 
richer $X$-conditional validity are an important direction for future work.

\begin{ack}
	We thank the anonymous reviewers for their valuable comments and constructive suggestions, which helped improve this work, particularly the clarity of the theoretical analysis and the experimental evaluation. This work was supported by the European Union's Horizon Europe
	research and innovation programme under the Marie
	Sk{\l}odowska-Curie COFUND grant agreement No.~101127936
	(DeMythif.AI), and by France 2030 funding, managed by the
	National Research Agency (ANR), as part of the IA CLUSTER
	programme, reference ANR-23-IACL-0003 -- DATAIA CLUSTER. The authors declare no competing interests.
\end{ack}

\bibliographystyle{plainnat} 
\bibliography{example_paper}

\clearpage
\appendix
\begin{center}
	\LARGE\bfseries Appendix
\end{center}
\section{Related Work}
\label{appendix:relatedwork}
\paragraph{Uncertainty prediction with missing values.} This is closely related to our work. Classic methods typically rely on impute-then-predict pipelines, where missing values are imputed with specific values and then standard uncertainty quantification tools, such as conformal prediction or quantile regression, are applied. Although such methods are valid for marginal coverage under very mild assumptions, more recent work, such as \citet{zaffran2023conformal, zaffran2024predictiveuncertaintyquantificationmissing}, has shown that the coverage and length of predictive intervals can vary significantly depending on the missingness pattern. They propose the missing data augmentation (MDA) technique to modify either the calibration dataset or the test point, ensuring that all points share the same mask to achieve MCV. However, their methods only provide theoretical guarantees for Missing Completely at Random (MCAR; i.e., $M \perp (X,Y)$), whereas in practice, real-world missing-data mechanisms are often more complex. 
Recent work \citep{fan2025arxiv} proposes a preimpute-then-mask framework and uses density-ratio correction to address a general missing mechanism.
The quality of the estimated density ratio affects statistical validity and interval efficiency, while computing the density-ratio correction introduces additional runtime overhead.
Other similar methods address this problem within the framework of matrix completion \citep{liang2024structured, gui2023conformalized, shao2023distribution}, proposing CP-based strategies with theoretical guarantees. Nevertheless, one of their basic assumptions is the random observation of columns, which is still MCAR, limiting the scope of their work to general adaptive uncertainty quantification under missing values.
Other similar work includes \citet{kong2026fair}, which proposes Fair CP and uses non-exchangeable (nexCP) and localized (LCP) conformal frameworks to achieve mask-conditional validity. They rely on an explicit definition of a distance metric on the space of masks.  \citet{tang2025conformal} use multiple candidate imputations for double calibration and reweighting non-conformity scores. Both methods provide theoretical guarantees on either MCAR or MAR and rely heavily on imputation. Instead, we learn an embedding of masked covariates by tree ensembles to handle the non-smooth geometry of the space of missing values.

\paragraph{Conformal prediction with conditional guarantees.}  Distribution-free pointwise $X$-conditional coverage is generally incompatible with informative sets without further assumptions about the data distributions \citep{vovk2012conditional}. In practice, there are two approaches to achieve a conditional guarantee. One direction, explored by \citet{gibbs2025conformal}, loosens the definition of conditional coverage—introducing approximate or relaxed versions that can be enforced without sacrificing informativeness. The conditional coverage by trust scores \citep{kaur2025conformal} can also be seen as a special case of this direction.  The second direction is through the estimations of the conditional distribution $Y|X$ or $S|X$. \citet{plassier2024probabilistic, lei2014distribution} estimated the conditional density and achieved asymptotic coverage, while  \citet{hore2025conformal, alaa2023conformalized} applied a localized approach that splits the feature space, followed by conformal prediction to achieve approximately conditional coverage. However, their methods typically assume the covariate space to be $\mbbR^d$, and, the estimation of  density is not trivial in practice. 

Our optimization and dual computation follow the pinball-loss framework of \citet{gibbs2025conformal}. However, the missing-data contribution is different: we construct a native representation specifically for masked covariates, augment it with mask indicators, and choose the penalty matrix to target different levels of validity. This is necessary because the masked covariate space lacks the geometric structure assumed by RKHS/Lipschitz conditional-coverage methods.

\section{Analysis of computational complexity}
\label{appendix:complex}
Let $n_{\mathrm{tr}},n_{\mathrm{cal}},n_{\mathrm{te}}$ denote the sample sizes for training, calibration, and test; $d$ denote the feature dimension; $P_{\mathrm{te}}$ and $P_{\mathrm{cal}}$ denote the numbers of distinct test and calibration missingness patterns; $p=n_{\tree}+d+1$ denote the dimension of the augmented feature map $\psi(x)=[\widehat T(x),\mathsf m(x),1]$; $T$ denote the number of trees used in the embedding; $G$ denote the number of grid points used by Weighted-CP; $I$ denote the number of bisection iterations for exact dual search; $S(n,p)$ denote the time of solving a convex LP/QP of size $(n,p)$, typically with $S(n,p)=O((n+p)^3)$ under interior-point solvers.

For each test point, we need $I = O(\log((S_{max}-S_{min})/tol))$ binary search to optimize. Therefore, we have the time complexity $
O(n_{\text{te}} \, I \, S(n_{\text{cal}},p)).
$
The dependence on the embedding dimension $p$ makes the exact dual path more expensive as the number of trees grows, while the empirical runtime study in Appendix~\ref{sec:runtime} shows that this cost remains manageable for the tree counts used in our experiments.

\section{Proofs}

\subsection{Proof of \Cref{thm:coord_mask_cond}}
\label{proof:thm1}

We begin from our goal:
\[
\mathbb{P}\left( Y^{n+1} \in \hat{C}_\alpha(X^{n+1}) \mid M^{n+1}_j = m_j \right)
\ge 1 - \alpha,
\]
which, by the definition of \Cref{eq:final-set} is equivalent to:
\begin{align*}
	& \mbbP ( Y^{n+1} \in \hat{C}_\alpha(X^{n+1}) \mid M^{n+1}_j = m_j )
	\ge 1 - \alpha \\
	\Leftrightarrow & \mbbP (	S\!(X^{n+1},Y^{n+1})
	\le
	\hat g_{S(X^{n+1},Y^{n+1})}(X^{n+1})  \mid M^{n+1}_j = m_j ) 	\ge 1 - \alpha \\
	\Leftrightarrow & \frac{\mathbb{E}[{\ind\{M^{n+1}_j =m_j\}}\ind\{S\!(X^{n+1},Y^{n+1})
		\le
		\hat g_{S(X^{n+1},Y^{n+1})}(X^{n+1}) \}]}{\mbbP(M^{n+1}_j =m_j)}  	\ge 1 - \alpha \\
	\Leftrightarrow & \frac{\mathbb{E}[{\ind\{M^{n+1}_j =m_j\}}\ind\{S\!(X^{n+1},Y^{n+1})
		\le
		\hat g_{S(X^{n+1},Y^{n+1})}(X^{n+1}) \}]}{\mbbP(M^{n+1}_j =m_j)}   	\\
        &\ge \frac{\mathbb{E}[{\ind\{M^{n+1}_j =m_j\}}(1-\alpha)]}{\mbbP(M^{n+1}_j =m_j)}  \\
	\Leftrightarrow & {\mathbb{E}[{\ind\{M^{n+1}_j =m_j\}}\ind\{S\!(X^{n+1},Y^{n+1})
		\le
		\hat g_{S(X^{n+1},Y^{n+1})}(X^{n+1}) \}]}	\\
        &\ge {\mathbb{E}[{\ind\{M^{n+1}_j =m_j\}}(1-\alpha)]} \\
\end{align*}
By denoting $S(X^{n+1},Y^{n+1})$ as $S^{n+1}$, our goal becomes 
\begin{equation}
	\label{eq:goal}
	{\mathbb{E}[{\ind\{M^{n+1}_j =m_j\}}\ind\{S^{n+1}
		\le
		\hat g_{S^{n+1}}(X^{n+1}) \}]}	\ge {\mathbb{E}[{\ind\{M^{n+1}_j =m_j\}}(1-\alpha)]} 
\end{equation}

Then from the other side, we check the first order condition of the minimizer Equation \eqref{eq:minimizer}:
\begin{align*}
	\hat{g}
	\;=\; \underset{g \in \mG}{\operatorname{argmin}} \frac{1}{|\mI_2|+1 }\sum_{i \in \mI_2 \cup \{n+1\}} \ell_{\alpha}\left(
	g\left(X^i\right), S^{i}\right) 
	\;+\;
	\lambda\mR(g) ,
\end{align*}
By the construction of $\mG$, we have $\ind\{M^{n+1}_j =m_j\} \in \mG$ for each $j$, since $M_j$ is explicitly included as a binary covariate in the model class.

Let us define the perturbation function:
\[
f(x) := {\ind\{M_j = m_j\}},
\]
which lies in $\mG$.

By convexity, the first-order optimality condition gives:
\begin{align}
	\label{eq:firstorder}
	0 \in \partial_\epsilon \left[
	\frac{1}{|\mI_2|+1 } \sum_{i \in \mI_2 \cup \{n+1\}} \ell_\alpha(\hat{g}_{S^{n+1}}(X^i) + \epsilon f(X^i), S^i)
	+ \lambda\mR(\hat{g}_{S^{n+1}} + \epsilon f)
	\right]_{\epsilon=0}. 
\end{align}
By the definition of $\mR$ which penalizes only the coefficients of tree-induced features, we have:
\begin{align}
	\mR(\hat{g}_{S^{n+1}} + \epsilon f) = \mR(\hat{g}_{S^{n+1}}) \Rightarrow   \frac{d}{d\epsilon} \mathcal{R}(\hat{g}_{S^{n+1}} + \epsilon f)=0 
\end{align}
Then the first-order condition becomes:
\begin{align}
	\label{eq:firstorder2}
	0 \in \partial_\epsilon \left[
	\frac{1}{|\mI_2|+1 } \sum_{i \in \mI_2 \cup \{n+1\}} \ell_\alpha(\hat{g}_{S^{n+1}}(X^i) + \epsilon f(X^i), S^i)
	\right]_{\epsilon=0}. 
\end{align}

We analyze the directional derivative of the pinball loss when $\epsilon \rightarrow0$ in three cases:

\begin{itemize}
	\item If \( S^i > \hat{g}_{S^{n+1}}(X^i) \), we take $\epsilon\leq\frac{S^i -\hat{g}_{S^{n+1}}(X^i)}{2}$ . Since $f(X^i)\leq1$ by definition, then the loss is 
	\[
	\ell_\alpha(\hat{g}_{S^{n+1}}(X^i) + \epsilon f(X^i), S^i)
	= (1 - \alpha)(S^i - \hat{g}_{S^{n+1}}(X^i) - \epsilon f(X^i)),
	\]
	and the directional derivative is
	\[
	\partial_\epsilon = - (1 - \alpha) f(X^i), \quad \forall \epsilon\leq\frac{S^i -\hat{g}_{S^{n+1}}(X^i)}{2}
	\]
	
	\item If \( S^i < \hat{g}_{S^{n+1}}(X^i) \), then we take $|\epsilon|\leq\frac{\hat{g}_{S^{n+1}}(X^i) - S^i}{2}$
	\[
	\ell_\alpha(\hat{g}_{S^{n+1}}(X^i) + \epsilon f(X^i), S^i)
	= \alpha(\hat{g}_{S^{n+1}}(X^i) + \epsilon f(X^i) - S^i),
	\]
	and the directional derivative is
	\[
	\partial_\epsilon = \alpha f(X^i), \quad \forall |\epsilon|\leq\frac{\hat{g}_{S^{n+1}}(X^i) - S^i}{2} .
	\]
	
	\item If \( S^i = \hat{g}_{S^{n+1}}(X^i) \), then the loss is non-differentiable. However, since it is convex, the subdifferential is the interval \( [\alpha - 1,\; \alpha] \), and there exists \( s_i^* \in [\alpha - 1, \alpha] \) such that
	\[
	\partial_\epsilon = s_i^* f(X^i).
	\]
\end{itemize}
By replacing the subgradients into \Cref{eq:firstorder2} where we consider the directional derivative when $\epsilon=0$, we obtain:
\begin{align}
	\label{eq:subgrad_full}
	0 = \frac{1}{|\mI_2| + 1} &(
	\sum_{i \in \mI_2 \cup \{n+1\} : S^i > \hat{g}_{S^{n+1}}(X^i)} - (1 - \alpha) f(X^i)
	+ \\
    &\sum_{i : S^i < \hat{g}_{S^{n+1}}(X^i)} \alpha f(X^i)
	+ \sum_{i : S^i = \hat{g}_{S^{n+1}}(X^i)} s_i^* f(X^i)
	), \nonumber
\end{align}
where each \( s_i^* \in [\alpha - 1, \alpha] \) is a valid subgradient for the nondifferentiable case \( S^i = \hat{g}_{S^{n+1}}(X^i) \).

We can equivalently write this expression as:
\begin{align}
	\label{eq:subgrad_final}
	\frac{1}{|\mI_2| + 1}
	\sum_{i \in \mI_2 \cup \{n+1\}} 
    &f(X^i)\cdot (
	\ind\{S^i \le \hat{g}_{S^{n+1}}(X^i)\}-(1-\alpha) \\
    &+ \ind\{S^i = \hat{g}_{S^{n+1}}(X^i)\} \cdot (s_i^* -\alpha )
	) = 0. \nonumber
\end{align}

Now, taking expectation over all randomness and under assumption of symmetrical optimization of $\hat{g}$, we obtain:
\begin{align}
	\mathbb{E}&[
	f(X^{n+1})
	\cdot \left(
	\ind\{S^{n+1} \le \hat{g}_{S^{n+1}}(X^{n+1})\}-(1-\alpha) \right) \nonumber \\
	&+ 
	\frac{\sum_{i \in \mI_2 \cup \{n+1\}} f(X^i)
		\cdot \left( \ind\{S^i = \hat{g}_{S^{n+1}}(X^i)\} \cdot (s_i^* -\alpha )
		\right) }{|\mI_2| + 1}	] 
	=0,
\end{align}
where \((s_i^* - \alpha ) \le 0 \) and $f(X^i) \ge 0$. By substituting $f(x) := {\ind\{M_j = m_j\}}$, we have 
\begin{align}
	\mathbb{E}[\ind\{M_j = m_j\}(
	\ind\{S^{n+1} \le \hat{g}_{S^{n+1}}(X^{n+1})\}-(1-\alpha))] \ge 0
\end{align}
which is exactly our goal in \Cref{eq:goal}.

\subsection{Proof of Proposition~\ref{prop:calib_mask_cond}}
\label{proof:observed-mask}
Fix a mask $m$ with $\pi_m>0$ and let $u=|\mathcal I_2|+1$. Construct the auxiliary procedure with indicators indexed by
\[
\mathcal M_{\mathrm{aug}}=\{M^i:i\in\mathcal I_2\cup\{n+1\}\},\qquad
N_m^{\mathrm{aug}}=\sum_{i\in\mathcal I_2\cup\{n+1\}}\mathbf1\{M^i=m\}.
\]
Both objects are invariant under permutations of all $u$ observations. Fit this procedure at the true test score using the symmetric selection rule, and put $H_i=\mathbf1\{S^i\le \hat g_{\mathrm{aug}}(X^i)\}-(1-\alpha)$. The same unpenalized-direction argument as in~\eqref{eq:subgrad_final}, including its nonpositive tie term, gives
\[
\sum_i\mathbf1\{M^i=m\}H_i\ge0.
\]
Multiply by the permutation-invariant event $E_m=\{N_m^{\mathrm{aug}}\ge2\}$ and take expectations. Exchangeability gives
\[
\mathbb E[\mathbf1\{M^{n+1}=m\}\mathbf1_{E_m}H_{n+1}]
=\frac1u\mathbb E\!\left[\mathbf1_{E_m}\sum_i\mathbf1\{M^i=m\}H_i\right]\ge0.
\]
Conditional on $M^{n+1}=m$, the event $E_m$ is exactly $\{m\in\mathcal M_{\mathrm{cal}}\}$. On that event $\mathcal M_{\mathrm{aug}}=\mathcal M_{\mathrm{cal}}$, so the objectives and selected minimizers coincide with the original procedure. Dividing by $\mathbb P(M^{n+1}=m,m\in\mathcal M_{\mathrm{cal}})>0$ proves the proposition.

\section{Generalization error}
\label{sec:generror}
\begin{definition}[Strongly Convex Functions]
A function $g:C \rightarrow \mbbR$ on a convex set
$C\subseteq\mbbR^p$ is $\gamma$-strongly convex ($\gamma>0$)
if $\forall x_1, x_2 \in C, \; \theta \in (0,1)$:
	\begin{equation}
		g(\theta x_1 + (1-\theta) x_2) \leq \theta g(x_1)+(1-\theta)g(x_2)-\frac{\gamma \theta (1-\theta)}{2}  \| x_1-x_2\|^2
	\end{equation}
\end{definition}

\begin{assumption}
	\label[assumption]{assump:strong-convex}
	The penalty matrix $J \in \mathbb{R}^{r \times (n_{\tree}+d+1)}$ satisfies
	\[
	\lambda_{\min}(J^\top J) = \lambda_J > 0, \quad	\lambda_{\max}(J^\top J) = \lambda_{J\mathrm{max}} < \infty, 
	\]
\end{assumption}

Based on \Cref{assump:strong-convex}, we can easily verify that the empirical risk function $\hat{L}(g)$ is $2 \lambda \lambda_J$-strongly convex.

\begin{definition}[Population risk and population minimizer]
	\label{def:populationrisk}
	The regularized population risk associated with $g_\beta \in \mathcal G$ is
	\[
	L(g_\beta) \;=\; 
	\mathbb{E}\!\left[\,\ell_\alpha\big(g_\beta(x), S\big)\,+ \lambda \, \mathcal R(g_\beta)\right],
	\]
	where the expectation is taken over the joint distribution of $(x, S)$.  
	We denote by 
	\[
	g^*_\beta \in \arg\min_{g_\beta \in \mathcal G_K} L(g_\beta)
	\]
	the population minimizer.
\end{definition}

\begin{lemma}
	For any fixed $s \in \mathbb{R}$, the pinball loss $\ell_\alpha(\cdot,s)$ is $C_\ell$-Lipschitz with constant $C_\ell = \max{\{\alpha,1-\alpha\}}$.
\end{lemma}

\begin{proof}
	We need to show that for all $\theta_1,\theta_2 \in \mathbb{R}$,
	\begin{equation*}
		\bigl|\ell_\alpha(\theta_1,s)-\ell_\alpha(\theta_2,s)\bigr|
		\le C_\ell\,|\theta_1-\theta_2|.
	\end{equation*}

	If both $\theta_1,\theta_2 \le s$, then
	\begin{equation*}
		\ell_\alpha(\theta,s) = (1-\alpha)(s-\theta),
	\end{equation*}
	
	so
	\begin{equation*}
		|\ell_\alpha(\theta_1,s)-\ell_\alpha(\theta_2,s)|
		= (1-\alpha)|\theta_1-\theta_2|
		\le C_\ell |\theta_1-\theta_2|.
	\end{equation*}
	
	If both $\theta_1,\theta_2 \ge s$, then
	\begin{equation*}
		\ell_\alpha(\theta,s) = \alpha(\theta-s),
	\end{equation*}
	and similarly,
	\begin{equation*}
		|\ell_\alpha(\theta_1,s)-\ell_\alpha(\theta_2,s)|
		= \alpha|\theta_1-\theta_2|
		\le C_\ell |\theta_1-\theta_2|.
	\end{equation*}
	
	Finally, if $\theta_1 \le s \le \theta_2$, then
	\begin{equation*}
		\ell_\alpha(\theta_1,s) = (1-\alpha)(s-\theta_1), \qquad
		\ell_\alpha(\theta_2,s) = \alpha(\theta_2-s).
	\end{equation*}
	Thus,
	\begin{align*}
		|\ell_\alpha(\theta_1,s)-\ell_\alpha(\theta_2,s)|
		&= \big|(1-\alpha)(s-\theta_1)-\alpha(\theta_2-s)\big| \\
		&\le (1-\alpha)|s-\theta_1| + \alpha|\theta_2-s| \\
		&\le C_\ell \big(|s-\theta_1| + |\theta_2-s|\big) \\
		&= C_\ell |\theta_2-\theta_1|,
	\end{align*}
	where the last equality holds because $s$ lies between $\theta_1$ and $\theta_2$.
	
	Combining the three cases establishes the claim.
\end{proof}

\begin{lemma}[Boundedness of the feature map]
	\label{lem:psi-bounded}
	Let $\psi(X) \in \mathbb{R}^p$ be the feature representation defined by
	\[
	\psi(X) = [\,\hat T(X),\, \mathsf m(X),\, 1\,],
	\]
	where $\hat T(X) \in \mathbb{R}^{n_{\tree}}$ is the tree embedding, 
	$\mathsf m(X) \in \{0,1\}^d$ is the missingness mask, and the last coordinate is a constant intercept.
	Then for all $X \in \mathcal{X}$, 
	\[
	\|\psi(X)\|_2 \le B_\psi,
	\]
	where one may take
	\[
	B_\psi = \sqrt{n_{\tree} C_T^2 + d + 1}, 
	\]
	and $C_T := \max_{t \in [n_{\tree}], \, \text{leaf}} |\,T^{(t)}(\cdot)\,|$ is the maximal absolute leaf value among all trees in the ensemble.
\end{lemma}

\begin{proof}
	By construction, each coordinate of $\hat T(X)$ is the leaf value of some decision tree, hence belongs to a finite set and is bounded in absolute value by $C_T$. Thus 
	$\|\hat T(X)\|_2^2 \le n_{\tree} C_T^2$. 
	The mask $\mathsf m(X)$ is a binary vector of dimension $d$, so $\|\mathsf m(X)\|_2^2 \le d$. 
	Finally, the intercept contributes $1$. Combining these bounds yields
	\[
	\|\psi(X)\|_2^2 \le n_{\tree} C_T^2 + d + 1,
	\]
	which proves the claim.
\end{proof}

\begin{proposition}
	\label{prop:stab-gen}
	Consider the regularized  \eqref{eq:minimizer} with hypothesis class $\mathcal G_K=\{g_\beta(X)=\langle\beta,\psi(X)\rangle, \|\beta\|_2 \le K\}$, where $\psi(X)=[\,\widehat T(X),\,\mathsf m(X),\,1\,]\in\mathbb R^{p}$, $p=n_{\tree}+d+1$. Under Assumption~\ref{assump:symmetric}  and \ref{assump:strong-convex}, let $g_{\hat\beta_{\mD}}=\arg\min_{g_\beta\in\mathcal G_K}\hat L_{\mD}(g_\beta)$ be the empirical minimizer on $\mD$. Then the expected generalization gap satisfies
	\[
	\mathbb{E}_{\mD}\!\left[\,L\!\left(g_{\hat\beta_{\mD}}\right)\;-\;\hat L_{\mD}\!\left(g_{\hat\beta_{\mD}}\right)\right]
	\;\le\;
	\frac{2\,C_\ell^{\,2}\,B_\psi^{\,2}}{\lambda\,\lambda_J}\cdot \frac{1}{\,|\mI_2|+1\,}.
	\]
	In particular, the rate is $O(1/n)$ with explicit constant $\displaystyle \frac{2\,C_\ell^{\,2}\,B_\psi^{\,2}}{\lambda\,\lambda_J}$. For the calibration-only sample, the same bound holds with $|\mI_2|$ in place of $|\mI_2|+1$.
\end{proposition}

\begin{proof}
	By definitions, we know that the dataset $\mathcal{D}:=\{(X^i, S^i)\}_{i \in \mI_2} \cup \{(X^{n+1}, S^{n+1})\}$ contains $|\mI_2|+1$ i.i.d points. We define an order of the points in $\mathcal{D}$ by the order of index in $\mI_2 \cup \{n+1\}$. 
	
	By reordering, we can rewrite $\mathcal{D}=\{(X^1,S^1), (X^2,S^2), ... (X^{|\mI_2| +1},S^{|\mI_2| +1})\}$.
	Let $(X',S')$ be an independent point of $(X, S)$.  For any $i\in\{1,\ldots,|\mI_2|+1\}$, we define the \emph{replace–one dataset}
	\begin{equation}
		\mathcal{D}^{(i)}:=\{(X^1,S^1), ..., (X^{i-1},S^{i-1}), (X',S'), (X^{i+1},S^{i+1}),...,(X^{|\mI_2|+1},S^{|\mI_2|+1})\}
	\end{equation}
	We define the empirical risk on a given dataset $\mathcal{D}$ is:
	\begin{equation*}
		\hat{L}_{\mathcal{D}}=\frac{1}{|\mI_2|+1} \sum_{(X,s) \in \mathcal{D}} \ell_{\alpha}\left(
		g_\beta\left(X\right), s\right)+
		\lambda \mR(g_\beta) ,
	\end{equation*}
	\begin{equation*}
		g_{\hat{\beta}_\mathcal{D}} = \arg\min_{g_\beta \in \mathcal G_K} \hat{L}_{\mathcal{D}}(g_\beta)
	\end{equation*}
	\begin{equation*}
		g_{\hat{\beta}_\mathcal{D}^{(i)}} = \arg\min_{g_\beta \in \mathcal G_K} \hat{L}_{\mathcal{D}^{(i)}}(g_\beta)
	\end{equation*}
	\begin{equation*}
		g_{{\beta}}^\star = \arg\min_{g_\beta \in \mathcal G_K} {L}(g_\beta)
	\end{equation*}
	
	Denote $p:=n_{\tree}+d+1$.
	Based on \Cref{assump:strong-convex}, for any $g \in \mG_K$, $\forall \theta \in (0,1)$
	\begin{align}
		&\hat{L}_{\mD}(g_{\hat{\beta}_\mathcal{D}} +\theta (g-g_{\hat{\beta}_\mathcal{D}}))  \leq \theta \hat{L}_{\mD}(g)+(1-\theta) \hat{L}_{\mD}(g_{\hat{\beta}_\mathcal{D}})-\frac{2\lambda \lambda_J \theta (1-\theta)}{2}\|g-g_{\hat{\beta}_\mathcal{D}}\|^2 \\
		\Rightarrow &\frac{\hat{L}_{\mD}(g_{\hat{\beta}_\mathcal{D}} +\theta (g-g_{\hat{\beta}_\mathcal{D}})) - 	\hat{L}_{\mD}(g_{\hat{\beta}_\mathcal{D}})}{\theta} \leq \hat{L}_{\mD}(g)-\hat{L}_{\mD}(g_{\hat{\beta}_\mathcal{D}})-\frac{2\lambda \lambda_J  (1-\theta)}{2}\|g-g_{\hat{\beta}_\mathcal{D}}\|^2 \\
		\Rightarrow &\lim_{\theta \rightarrow 0} \frac{\hat{L}_{\mD}(g_{\hat{\beta}_\mathcal{D}} +\theta (g-g_{\hat{\beta}_\mathcal{D}})) - 	\hat{L}_{\mD}(g_{\hat{\beta}_\mathcal{D}})}{\theta} \leq \lim_{\theta \rightarrow 0} \hat{L}_{\mD}(g)-\hat{L}_{\mD}(g_{\hat{\beta}_\mathcal{D}})-\frac{2\lambda \lambda_J  (1-\theta)}{2}\|g-g_{\hat{\beta}_\mathcal{D}}\|^2 \\
		\Rightarrow &0 \leq  \hat{L}_{\mD}(g)-\hat{L}_{\mD}(g_{\hat{\beta}_\mathcal{D}})-{\lambda \lambda_J }\|g-g_{\hat{\beta}_\mathcal{D}}\|^2
	\end{align}
	We take $g=g_{\hat{\beta}_\mathcal{D}^{(i)}} $ and we have:
	\begin{align}
		& 0 \leq  \hat{L}_{\mD}(g_{\hat{\beta}_\mathcal{D}^{(i)}} )-\hat{L}_{\mD}(g_{\hat{\beta}_\mathcal{D}})-{\lambda \lambda_J }\|g_{\hat{\beta}_\mathcal{D}^{(i)}} -g_{\hat{\beta}_\mathcal{D}}\|^2 \\
		\Leftrightarrow & {\lambda \lambda_J }\|g_{\hat{\beta}_\mathcal{D}^{(i)}} -g_{\hat{\beta}_\mathcal{D}}\|^2 \leq  \hat{L}_{\mD}(g_{\hat{\beta}_\mathcal{D}^{(i)}} )-\hat{L}_{\mD}(g_{\hat{\beta}_\mathcal{D}}) \label{eq:sc-used} 
	\end{align}
	On the other hand, the regularizer is the same in $\hat L_{\mD}$ and $\hat L_{\mD^{(i)}}$ for a fixed function, so the replace-one decomposition only changes the two loss evaluations at $(X^i,s^i)$ and $(X',s')$:
	\begin{align}
		&({|\mI_2|+1})(\hat{L}_{\mD}(g_{\hat{\beta}_\mathcal{D}^{(i)}} )-\hat{L}_{\mD}(g_{\hat{\beta}_\mathcal{D}}))\\
		=&
		 \sum_{(X,s) \in \mathcal{D}} \ell_{\alpha}\left(
		g_{\hat{\beta}_\mathcal{D}^{(i)}}\left(X\right), s\right) -\sum_{(X,s) \in \mathcal{D}} \ell_{\alpha}\left(
		g_{\hat{\beta}_\mathcal{D}}\left(X\right), s\right)+
		(|\mI_2|+1)\lambda (\mR(g_{\hat{\beta}_\mathcal{D}^{(i)}})
-\mR(g_{\hat{\beta}_\mathcal{D}}))  \\
		=&\sum_{(X,s) \in \mathcal{D}^{(i)}} \ell_{\alpha}\left(
		g_{\hat{\beta}_\mathcal{D}^{(i)}}\left(X\right), s\right) -\sum_{(X,s) \in \mathcal{D}^{(i)}} \ell_{\alpha}\left(
		g_{\hat{\beta}_\mathcal{D}}\left(X\right), s\right)+
		(|\mI_2|+1)\lambda (\mR(g_{\hat{\beta}_\mathcal{D}^{(i)}})
-\mR(g_{\hat{\beta}_\mathcal{D}})) \nonumber \\
		&- \ell_{\alpha}(g_{\hat{\beta}_\mathcal{D}^{(i)}}(X'),s')
		+ \ell_{\alpha}(g_{\hat{\beta}_\mathcal{D}^{(i)}}(X^i),s^i)
		+ \ell_{\alpha}(g_{\hat{\beta}_\mathcal{D}}(X'),s')- \ell_{\alpha}(g_{\hat{\beta}_\mathcal{D}}(X^i),s^i) \\
		=&(|\mI_2|+1)\Bigl(
    \hat{L}_{\mD^{(i)}}(g_{\hat{\beta}_\mathcal{D}^{(i)}})
    -\hat{L}_{\mD^{(i)}}(g_{\hat{\beta}_\mathcal{D}})
    \Bigr) \nonumber \\
        &- \ell_{\alpha}(g_{\hat{\beta}_\mathcal{D}^{(i)}}(X'),s') 
		+ \ell_{\alpha}(g_{\hat{\beta}_\mathcal{D}^{(i)}}(X^i),s^i)
		+ \ell_{\alpha}(g_{\hat{\beta}_\mathcal{D}}(X'),s')- \ell_{\alpha}(g_{\hat{\beta}_\mathcal{D}}(X^i),s^i) \\
		\leq &  0- \ell_{\alpha}(g_{\hat{\beta}_\mathcal{D}^{(i)}}(X'),s')
		+ \ell_{\alpha}(g_{\hat{\beta}_\mathcal{D}^{(i)}}(X^i),s^i)
		+ \ell_{\alpha}(g_{\hat{\beta}_\mathcal{D}}(X'),s')- \ell_{\alpha}(g_{\hat{\beta}_\mathcal{D}}(X^i),s^i) \\
		= & 
		\underbrace{\ell_\alpha(g_{\hat\beta_{\mD}^{(i)}}(X^i),s^i)-\ell_\alpha(g_{\hat\beta_{\mD}}(X^i),s^i)}_{A_i}
		+
		\underbrace{\ell_\alpha(g_{\hat\beta_{\mD}}(X'),s')-\ell_\alpha(g_{\hat\beta_{\mD}^{(i)}}(X'),s')}_{B_i},
		\label{eq:split}
	\end{align}
	
	where we used the optimality of $g_{\hat\beta_{\mD}^{(i)}}$ for the dataset $\mD^{(i)}$ to conclude 
	$\hat L_{\mD^{(i)}}(g_{\hat\beta_{\mD}^{(i)}})-\hat L_{\mD^{(i)}}(g_{\hat\beta_{\mD}})\le 0$.

	We have shown that $\|\psi(X)\|_2\le B_\psi$  and recall that 
	$\ell_\alpha(\cdot,s)$ is $C_\ell$-Lipschitz in its first argument, 
	$C_\ell=\max\{\alpha,1-\alpha\}$. Then
    \begin{align*}
        |A_i|
	&=
	\big|\ell_\alpha(g_{\hat\beta_{\mD}^{(i)}}(X^i),s^i)-\ell_\alpha(g_{\hat\beta_{\mD}}(X^i),s^i)\big|
	\le C_\ell\,\big|g_{\hat\beta_{\mD}^{(i)}}(X^i)-g_{\hat\beta_{\mD}}(X^i)\big| \\
	&= C_\ell\,\big|\langle\psi(X^i),\,\hat\beta_{\mD}^{(i)}-\hat\beta_{\mD}\rangle\big|
	\le C_\ell B_\psi\,\|\hat\beta_{\mD}^{(i)}-\hat\beta_{\mD}\|.
    \end{align*}
	Similarly,
	\[
	|B_i|
	=
	\big|\ell_\alpha(g_{\hat\beta_{\mD}}(X'),s')-\ell_\alpha(g_{\hat\beta_{\mD}^{(i)}}(X'),s')\big|
	\le C_\ell\,\big|g_{\hat\beta_{\mD}}(X')-g_{\hat\beta_{\mD}^{(i)}}(X')\big|
	\le C_\ell B_\psi\,\|\hat\beta_{\mD}^{(i)}-\hat\beta_{\mD}\|.
	\]
	Plugging these into \eqref{eq:split}, we obtain
	\begin{equation}\label{eq:split-bounded}
		\hat L_{\mD}(g_{\hat\beta_{\mD}^{(i)}})-\hat L_{\mD}(g_{\hat\beta_{\mD}})
		\;\le\; \frac{2C_\ell B_\psi}{|\mI_2|+1}\,\|\hat\beta_{\mD}^{(i)}-\hat\beta_{\mD}\|.
	\end{equation}
	
	Combining \eqref{eq:sc-used} and \eqref{eq:split-bounded} gives
	\[
	\lambda\lambda_J\,\|\hat\beta_{\mD}^{(i)}-\hat\beta_{\mD}\|^2
	\;\le\; \frac{2C_\ell B_\psi}{|\mI_2|+1}\,\|\hat\beta_{\mD}^{(i)}-\hat\beta_{\mD}\|.
	\]
	If $\hat\beta_{\mD}^{(i)}\ne\hat\beta_{\mD}$, divide both sides by 
	$\|\hat\beta_{\mD}^{(i)}-\hat\beta_{\mD}\|$ to obtain (Otherwise the bound is trivial.)
	
	\begin{equation}\label{eq:beta-diff}
		\|\hat\beta_{\mD}^{(i)}-\hat\beta_{\mD}\|
		\;\le\; \frac{2C_\ell B_\psi}{\lambda\lambda_J}\cdot\frac{1}{|\mI_2|+1}.
	\end{equation}
	
	For any test point $(X,s)$,
    \begin{align*}
        \big|\ell_\alpha(g_{\hat\beta_{\mD}}(X),s)-\ell_\alpha(g_{\hat\beta_{\mD}^{(i)}}(X),s)\big|
	&\le C_\ell\,\big|g_{\hat\beta_{\mD}}(X)-g_{\hat\beta_{\mD}^{(i)}}(X)\big|\\
    &= C_\ell\,\big|\langle\psi(X),\,\hat\beta_{\mD}-\hat\beta_{\mD}^{(i)}\rangle\big|
	\le C_\ell B_\psi\,\|\hat\beta_{\mD}-\hat\beta_{\mD}^{(i)}\|.
    \end{align*}
	Using \eqref{eq:beta-diff} we conclude the uniform stability bound
	\begin{equation}\label{eq:beta-stability}
		\big|\ell_\alpha(g_{\hat\beta_{\mD}}(X),s)-\ell_\alpha(g_{\hat\beta_{\mD}^{(i)}}(X),s)\big|
		\;\le\; \frac{2C_\ell^2 B_\psi^2}{\lambda\lambda_J}\cdot\frac{1}{|\mI_2|+1}.
	\end{equation}
	
	Let $i$ be uniform on $\{1,\ldots,|\mI_2|+1\}$ and independent of $(\mD,(X',S'))$.
	By exchangeability of the sample and permutation invariance of the learning rule,
	\[
	\begin{aligned}
		\mathbb{E}_{\mD}\big[L(g_{\hat\beta_{\mD}})\big]
		&= \mathbb{E}_{\mD,(X',S')}\big[\ell_\alpha(g_{\hat\beta_{\mD}}(X'),S')\big]
		+\lambda\mathbb{E}_{\mD}\big[\mR(g_{\hat\beta_{\mD}})\big]\\
		&= \mathbb{E}_{\mD,(X',S'),i}\big[\ell_\alpha(g_{\hat\beta_{\mD^{(i)}}}(X^i),S^i)\big]
		+\lambda\mathbb{E}_{\mD}\big[\mR(g_{\hat\beta_{\mD}})\big],
	\end{aligned}
	\]
	and
	\[
	\mathbb{E}_{\mD}\big[\hat L_{\mD}(g_{\hat\beta_{\mD}})\big]
	= \mathbb{E}_{\mD,i}\big[\ell_\alpha(g_{\hat\beta_{\mD}}(X^i),S^i)\big]
	+\lambda\mathbb{E}_{\mD}\big[\mR(g_{\hat\beta_{\mD}})\big].
	\]
	Subtracting the two identities, the regularization terms cancel.
	Therefore,
    \begin{align*}
        \mathbb{E}_{\mD}\!\left[L(g_{\hat\beta_{\mD}})-\hat L_{\mD}(g_{\hat\beta_{\mD}})\right]
	&= \mathbb{E}_{\mD,(X',S'),i}\!\left[\ell_\alpha(g_{\hat\beta_{\mD^{(i)}}}(X^i),S^i)-\ell_\alpha(g_{\hat\beta_{\mD}}(X^i),S^i)\right] \\
    &\le \frac{2C_\ell^2 B_\psi^2}{\lambda\lambda_J}\cdot\frac{1}{|\mI_2|+1},
    \end{align*}
	where the last inequality uses \eqref{eq:beta-stability} with $(X,s)=(X^i,S^i)$.
\end{proof}

\begin{assumption}[Bounded scores and bounded predictions]
	\label[assumption]{assump:bounded-scores}
	There exist constants $S_{\max}>0$ such that for all $(x,s)$ in the domain,
	the nonconformity score satisfies $|s|\le S_{\max}$.
\end{assumption}

\begin{remark}
	For the pinball loss, which is $C_\ell$-Lipschitz in its first argument with
	$C_\ell=\max\{\alpha,1-\alpha\}$, Assumption~\ref{assump:bounded-scores} implies a uniform loss bound:
	\[
	0 \le \ell_\alpha\big(g_\beta(x),s\big) 
	\le C_\ell\big(|g_\beta(x)|+|s|\big) 
	\le B_\ell
	\quad\text{with}\quad B_\ell:=C_\ell\,(KB_\psi+S_{\max}).
	\]
	Hence Assumption~\ref{assump:bounded-scores} ensures bounded loss.
\end{remark}

\begin{proposition}[PAC-like generalization bound]
	\label{prop:pac-like}
	Assume the conditions of Proposition~\ref{prop:stab-gen} 
	and additionally Assumption~\ref{assump:bounded-scores}.
	Let 
	\[
	\beta_u \;:=\; \frac{2\,C_\ell^{\,2}\,B_\psi^{\,2}}{\lambda\,\lambda_J}\cdot \frac{1}{u},
	\qquad u:=|\mathcal D|,
	\]
	be the uniform stability constant established in~\eqref{eq:beta-stability}. 
	Then for any $\delta\in(0,1)$, with probability at least $1-\delta$ over the draw of $\mathcal D$,
	\[
	L\!\big(g_{\hat\beta_{\mathcal D}}\big)\;-\;\hat L_{\mathcal D}\!\big(g_{\hat\beta_{\mathcal D}}\big)
	\;\le\;
	\beta_u \;+\; \big(4u\,\beta_u + B_\ell\big)\,
	\sqrt{\frac{\ln(1/\delta)}{2u}},
	\]
\end{proposition}

\begin{proof}
	Define $f(\mathcal D):=L(g_{\hat\beta_{\mathcal D}})-\hat L_{\mathcal D}(g_{\hat\beta_{\mathcal D}})$. Define
	$L_0(g):=\mathbb E[\ell_\alpha(g(X),S)]$ and
	$\hat L_{0,\mathcal D}(g):=u^{-1}\sum_{(X,s)\in\mathcal D}\ell_\alpha(g(X),s)$, such that $\hat L_{\mathcal D}(g)=\hat L_{0,\mathcal D}(g)+\lambda\mathcal R(g)
	$.
	Because the same regularizer appears in both $L$ and $\hat L_{\mathcal D}$, we have also:
	\[
	f(\mathcal D)=L_0(g_{\hat\beta_{\mathcal D}})
	-\hat L_{0,\mathcal D}(g_{\hat\beta_{\mathcal D}}).
	\]
	Thus, the bounded-difference argument below only involves the unregularized pinball loss.
	By Proposition~\ref{prop:stab-gen} (expected bound), $\mathbb E_{\mathcal D}[f(\mathcal D)]\le \beta_u$.
	Fix $i\in\{1,\ldots,u\}$ and form the replace–one dataset 
	$\mathcal D^{(i)}$ by substituting the $i$-th point of $\mathcal D$ with an independent copy
	(as defined in the proof preceding~\eqref{eq:split}).
	We show that $f$ is a bounded–difference function w.r.t.\ the $i$-th coordinate:
	\[
	\big|f(\mathcal D)-f(\mathcal D^{(i)})\big|
	\;\le\; 4\,\beta_u \;+\; \frac{B_\ell}{u}.
	\]
	Indeed, write
	\[
	f(\mathcal D)-f(\mathcal D^{(i)})
	=\underbrace{L_0(g_{\hat\beta_{\mathcal D}})-L_0(g_{\hat\beta_{\mathcal D^{(i)}}})}_{\mathrm{(A)}}
	\;-\;
	\underbrace{\big(\hat L_{0,\mathcal D}(g_{\hat\beta_{\mathcal D}})-\hat L_{0,\mathcal D^{(i)}}(g_{\hat\beta_{\mathcal D^{(i)}}})\big)}_{\mathrm{(B)}}.
	\]
	For $\mathrm{(A)}$, by uniform stability (see~\eqref{eq:beta-stability}), changing the dataset by one point changes the loss at any test pair by at most $\beta_u$, hence
	$|\,\mathrm{(A)}\,|\le\beta_u\le2\beta_u$, retaining a loose constant for consistency with the displayed bound.
	For $\mathrm{(B)}$, decompose
	\[
	\hat L_{0,\mathcal D}(g_{\hat\beta_{\mathcal D}})-\hat L_{0,\mathcal D^{(i)}}(g_{\hat\beta_{\mathcal D^{(i)}}})
	=
	\big(\hat L_{0,\mathcal D}(g_{\hat\beta_{\mathcal D}})-\hat L_{0,\mathcal D}(g_{\hat\beta_{\mathcal D^{(i)}}})\big)
	+\big(\hat L_{0,\mathcal D}(g_{\hat\beta_{\mathcal D^{(i)}}})-\hat L_{0,\mathcal D^{(i)}}(g_{\hat\beta_{\mathcal D^{(i)}}})\big).
	\]
	The first difference is bounded by $\beta_u\le2\beta_u$ via stability, since each of the $u$ per-point loss differences is at most $\beta_u$.
	The second difference is an average that only differs in the $i$-th term, hence its absolute value is bounded by $B_\ell/u$ by loss boundedness. Altogether,
	$|\,\mathrm{(B)}\,|\le 2\beta_u + B_\ell/u$, whence
	\[
	|f(\mathcal D)-f(\mathcal D^{(i)})|\le 2\beta_u + (2\beta_u+B_\ell/u) = 4\beta_u + B_\ell/u.
	\]
	
	By McDiarmid's inequality for independent samples and bounded differences,
	\[
	\mathbb P\!\left(f(\mathcal D)-\mathbb E f(\mathcal D) > t\right)
	\;\le\;
	\exp\!\Big(-\tfrac{2u\,t^2}{(4u\beta_u+B_\ell)^2}\Big).
	\]
	Setting the right-hand side equal to $\delta$ gives
	$t=(4u\beta_u+B_\ell)\sqrt{\ln(1/\delta)/(2u)}$.
	Since $\mathbb E f(\mathcal D)\le \beta_u$, we obtain, with probability at least $1-\delta$,
	\[
	f(\mathcal D)\le \beta_u + (4u\beta_u+B_\ell)\sqrt{\frac{\ln(1/\delta)}{2u}},
	\]
	which is the desired bound.
\end{proof}

\begin{assumption}
	\label[assumption]{assump:gram}
	The feature Gram matrix $G=\mathbb E[\psi(X)\psi(X)^\top]$ satisfies
	$0<\lambda_{\min}(G)\le \lambda_{\max}(G)<\infty$.
\end{assumption}

\begin{theorem}
	\label{thm:mask-cond-generalJ_app}
	Suppose Assumptions~\ref{assump:linearind}, \ref{assump:symmetric}, \ref{assump:mpropa}, \ref{assump:anti-concentration}, \ref{assump:strong-convex}, \ref{assump:bounded-scores}, and \ref{assump:gram} hold, with $\lambda>0$ and $0<\alpha<1$. Let $\mathcal C$ be the calibration sample, write $n=|\mathcal I_2|\ge1$ locally in this proof, and set $u=n+1$. Define
	\[
	F_{\mathcal C}(\beta)=\frac1n\sum_{i\in\mathcal I_2}\ell_\alpha(g_\beta(X^i),S^i)+\lambda\|J\beta\|_2^2,
	\qquad
	\widetilde\beta_{\mathcal C}=\arg\min_{\beta\in B_K}F_{\mathcal C}(\beta).
	\]
	Let $\widehat\beta_{\mathcal C,Z}$ minimize the augmented objective~\eqref{eq:regularised loss} over $B_K$, and let $\beta^\star$ minimize $L$ over $B_K$. Assume $\kappa=K-\|\beta^\star\|_2>0$. For the normalized indicator $r_m=\mathbf1\{M=m\}/\pi_m$, write $r_K=g_{\eta_{K,m}}$, $\|\eta_{K,m}\|_2\le K$, for its projection onto $\mathcal G_K$ and $\Delta_{K,m}=\|r_m-r_K\|_{L^2(\mathbb P_X)}$.
	Set
	\begin{align}
		b_n&=\frac{2C_\ell^2B_\psi^2}{\lambda\lambda_J n},\\
		e_n(\delta)&=b_n+(4nb_n+B_\ell)\sqrt{\frac{\log(1/\delta)}{2n}},\\
		\varepsilon_{\mathrm{cal}}(\delta)&=e_n(\delta/2)+B_\ell\sqrt{\frac{\log(2/\delta)}{2n}}, \label{eq:cal-excess-def} \\
		L_{\mathrm{sm}}&=C_0B_\psi^2+2\lambda\lambda_{J\mathrm{max}}.
	\end{align}
	Define
	\begin{align}
		D_n&=\frac{4K^2}{\kappa}\sqrt{2L_{\mathrm{sm}}\varepsilon_{\mathrm{cal}}(\delta)}
		+\frac{4K}{\kappa}\varepsilon_{\mathrm{cal}}(\delta),\\
		R_\lambda&=2K\sqrt{\lambda\lambda_{J\mathrm{max}}B_\ell},\\
		\tau_n&=\min\left\{B_\psi\sqrt{\frac{2B_\ell}{\lambda\lambda_J(n+1)}},
		\frac{C_\ell B_\psi^2}{\lambda\lambda_J(n+1)}\right\},\\
		T_n&=KB_\psi\min\{1,\bar C_0\tau_n\},\label{eq:repair-constants}
	\end{align}
	where $\bar C_0=\max\{C_0,1/t_0\}$. Then, with probability at least $1-\delta$ over $\mathcal C$ alone,
	\begin{equation}\label{eq:mcov-thm}
		\mathbb P\!\left(Y\in\widehat C_\alpha(X)\,\middle|\,M=m,\mathcal C\right)
		\ge1-\alpha-D_n-R_\lambda-T_n-c_\alpha\Delta_{K,m},
		\qquad c_\alpha=\max\{\alpha,1-\alpha\}.
	\end{equation}
\end{theorem}

\begin{proof}
	The test pair $Z=(X,S(X,Y))$ is independent of $\mathcal C$ but is included in $\widehat\beta_{\mathcal C,Z}$. Thus $\{S\le g_{\widehat\beta_{\mathcal C,Z}}(X)\}$ is precisely the inclusion event in~\eqref{eq:final-set}. We abbreviate $L(g_\beta)$ as $L(\beta)$.
	
	\medskip
	\noindent\emph{Step 1: Population smoothness for a fixed coefficient vector.}
	Under \cref{assump:anti-concentration},
	\[
	\mathbb P(|S-g_\beta(X)|\le t)\le C_0t\longrightarrow0
	\quad(t\downarrow0),
	\]
	so $\mathbb P(S=g_\beta(X))=0$ for every fixed $\beta\in B_K$. Bounded pinball subgradients and bounded features justify differentiation under the expectation:
	\begin{equation}\label{eq:nabla-pop}
		\nabla L(\beta)=
		\mathbb E\!\left[\bigl(\mathbf1\{S\le g_\beta(X)\}-(1-\alpha)\bigr)\psi(X)\right]
		+2\lambda J^\top J\beta.
	\end{equation}
	When $B_\psi\|\beta-\beta'\|_2\le t_0$,
	\begin{align*}
		&\|\nabla L(\beta)-\nabla L(\beta')\|_2 \\
		\le &B_\psi\mathbb P\!\left(|S-g_\beta(X)|\le B_\psi\|\beta-\beta'\|_2\right)
		+2\lambda\lambda_{J\mathrm{max}}\|\beta-\beta'\|_2\\
		\le & (C_0B_\psi^2+2\lambda\lambda_{J\mathrm{max}})\|\beta-\beta'\|_2.
	\end{align*}
	For any two points in $B_K$, subdivide the segment joining them into sufficiently short pieces and sum this inequality. Convexity of $B_K$ keeps each piece feasible, proving that $\nabla L$ is $L_{\mathrm{sm}}$-Lipschitz on the whole ball. The interior minimizer satisfies $\nabla L(\beta^\star)=0$.

	By using the Lipschitz condition of $\nabla L$, we have for any $x,y\in B_K$
	$$L(y)
	\le
	L(x)+\langle \nabla L(x),y-x\rangle
	+
	\frac{	L_{\mathrm{sm}}}{2}\|y-x\|_2^2.$$
	
	\medskip
	\noindent\emph{Step 2: Calibration-only excess risk and a feasible-direction gradient bound.}
	Apply Proposition~\ref{prop:pac-like} to the $n$ calibration points at level $\delta/2$, and Hoeffding's inequality to the fixed $\beta^\star$ at level $\delta/2$. By a union bound, with probability at least $1-\delta$ over $\mathcal C$,
	\begin{align}
		L(\widetilde\beta_{\mathcal C})-L(\beta^\star)
		={}&\underbrace{F_{\mathcal C}(\widetilde\beta_{\mathcal C})-F_{\mathcal C}(\beta^\star)}_{\le0}
		+\underbrace{L(\widetilde\beta_{\mathcal C})-F_{\mathcal C}(\widetilde\beta_{\mathcal C})}_{\le e_n(\delta/2)}
		+\underbrace{F_{\mathcal C}(\beta^\star)-L(\beta^\star)}_{\le B_\ell\sqrt{\log(2/\delta)/(2n)}}\nonumber\\
		\le{}&\varepsilon_{\mathrm{cal}}(\delta).\label{eq:cal-excess}
	\end{align}
	To control the gradient using feasible directions, for any $x\in B_K$ with $L(x)-L(\beta^\star)\le\varepsilon$ and nonzero $\eta$, let $v=\eta/\|\eta\|_2$, $a=\kappa/2$, and $w_\pm=\beta^\star\pm av$. Both $w_\pm$ and $y_\pm(t)=(1-t)x+tw_\pm$ lie in $B_K$ for $0\le t\le1$. Smoothness and minimality of $\beta^\star$ give
	\[
	L(\beta^\star)\le L(y_\pm(t))
	\le L(x)+t\langle\nabla L(x),\beta^\star-x\pm av\rangle
	+\frac{L_{\mathrm{sm}}t^2}{2}\|w_\pm-x\|_2^2.
	\]
	Writing $c=\langle\nabla L(x),\beta^\star-x\rangle\le L(\beta^\star)-L(x)\le0$, $q=\langle\nabla L(x),v\rangle$, and using $\|w_\pm-x\|_2\le2K$, we obtain
	\[
	c\pm aq\ge-\frac{\varepsilon}{t}-2L_{\mathrm{sm}}K^2t=:-R_t.
	\]
	Hence $|aq|\le R_t+c\le R_t$. For $\varepsilon>0$, choose
	$t=\min\{1,\sqrt{\varepsilon/(2L_{\mathrm{sm}}K^2)}\}$ to get
	$R_t\le2K\sqrt{2L_{\mathrm{sm}}\varepsilon}+2\varepsilon$; for $\varepsilon=0$, let $t\downarrow0$. Therefore
	\begin{equation}\label{eq:grad-gap}
		|\langle\eta,\nabla L(x)\rangle|
		\le\frac{4K\|\eta\|_2}{\kappa}\sqrt{2L_{\mathrm{sm}}\varepsilon}
		+\frac{4\|\eta\|_2}{\kappa}\varepsilon.
	\end{equation}
	The case $\eta=0$ is immediate. Substituting $x=\widetilde\beta_{\mathcal C}$, $\eta=\eta_{K,m}$, and~\eqref{eq:cal-excess} yields $|\langle\eta_{K,m},\nabla L(\widetilde\beta_{\mathcal C})\rangle|\le D_n$.
	
	\medskip
	\noindent\emph{Step 3: Regularization and the anchor's projected coverage moment.}
	Conditional on $\mathcal C$, the anchor $\widetilde\beta_{\mathcal C}$ is fixed and the test pair is fresh, so~\eqref{eq:nabla-pop} gives
	\begin{equation}\label{eq:dir-id}
		\mathbb E[r_K(X)h^0_{\mathcal C}(Z)\mid\mathcal C]
		=\langle\eta_{K,m},\nabla L(\widetilde\beta_{\mathcal C})\rangle
		-2\lambda\langle J\eta_{K,m},J\widetilde\beta_{\mathcal C}\rangle,
	\end{equation}
	where $h^0_{\mathcal C}(Z)=\mathbf1\{S\le g_{\widetilde\beta_{\mathcal C}}(X)\}-(1-\alpha)$.
	Since $\beta=0$ is feasible and the loss is nonnegative,
	\[
	\lambda\|J\widetilde\beta_{\mathcal C}\|_2^2
	\le F_{\mathcal C}(\widetilde\beta_{\mathcal C})
	\le F_{\mathcal C}(0)\le B_\ell.
	\]
	By Cauchy--Schwarz and $\|J\eta_{K,m}\|_2\le\sqrt{\lambda_{J\mathrm{max}}}K$,
	\[
	2\lambda|\langle J\eta_{K,m},J\widetilde\beta_{\mathcal C}\rangle|
	\le2K\sqrt{\lambda\lambda_{J\mathrm{max}}B_\ell}=R_\lambda.
	\]
	Consequently,
	\begin{equation}\label{eq:rhoK-cover}
		\mathbb E[r_Kh^0_{\mathcal C}\mid\mathcal C]\ge-D_n-R_\lambda.
	\end{equation}
	
	\medskip
	\noindent\emph{Step 4: Deterministic add-one stability.}
	For any candidate pair $z$, let $F_{\mathcal C,z}$ be the augmented objective and $d_z=\widehat\beta_{\mathcal C,z}-\widetilde\beta_{\mathcal C}$. Both objectives are $\mu=2\lambda\lambda_J$-strongly convex on the same ball. Since $0\le\ell_z(\beta)\le B_\ell$,
	\[
	\sup_{\beta\in B_K}|F_{\mathcal C,z}(\beta)-F_{\mathcal C}(\beta)|
	=\sup_{\beta\in B_K}\frac{|\ell_z(\beta)-n^{-1}\sum_i\ell_i(\beta)|}{n+1}
	\le\frac{B_\ell}{n+1}.
	\]
	Optimality and strong convexity therefore imply
	\[
	\frac{\mu}{2}\|d_z\|_2^2
	\le F_{\mathcal C}(\widehat\beta_{\mathcal C,z})-F_{\mathcal C}(\widetilde\beta_{\mathcal C})
	\le\frac{2B_\ell}{n+1},
	\]
	which gives the first term in $\tau_n$ after multiplication by $B_\psi$.
	For the sharper bound, put $A_z=F_{\mathcal C,z}-F_{\mathcal C}$. Its Lipschitz constant in $\beta$ is at most $2C_\ell B_\psi/(n+1)$. Adding the two strong-convexity inequalities at the respective minimizers gives
	\[
	\mu\|d_z\|_2^2
	\le A_z(\widetilde\beta_{\mathcal C})-A_z(\widehat\beta_{\mathcal C,z})
	\le\frac{2C_\ell B_\psi}{n+1}\|d_z\|_2.
	\]
	Thus
	\begin{equation}\label{eq:add-one}
		\sup_z\sup_x|g_{\widehat\beta_{\mathcal C,z}}(x)-g_{\widetilde\beta_{\mathcal C}}(x)|
		\le\tau_n.
	\end{equation}
	The supremum is over candidate pairs in the score support; the sharper Lipschitz bound also holds for arbitrary candidate scores without boundedness.
	
	\medskip
	\noindent\emph{Step 5: Transfer the projected moment, then control the residual.}
	Define $h^{\mathrm{aug}}_{\mathcal C}(Z)=\mathbf1\{S\le g_{\widehat\beta_{\mathcal C,Z}}(X)\}-(1-\alpha)$. The exact decomposition is
	\begin{align}
		\mathbb E[r_mh^{\mathrm{aug}}_{\mathcal C}\mid\mathcal C]
		={}&\mathbb E[r_Kh^0_{\mathcal C}\mid\mathcal C]
		+\mathbb E[r_K(h^{\mathrm{aug}}_{\mathcal C}-h^0_{\mathcal C})\mid\mathcal C]+\mathbb E[(r_m-r_K)h^{\mathrm{aug}}_{\mathcal C}\mid\mathcal C].\label{eq:projected-transfer}
	\end{align}
	By~\eqref{eq:add-one}, the two indicators can differ only if
	$|S-g_{\widetilde\beta_{\mathcal C}}(X)|\le\tau_n$. Assumption~\ref{assump:anti-concentration} gives, for all $t>0$,
	\[
	\mathbb P(|S-g_\beta(X)|\le t)\le\min\{1,\bar C_0t\}:
	\]
	for $t\le t_0$ use the assumption, and for $t>t_0$ use $\bar C_0t\ge1$. Since $|r_K(X)|\le KB_\psi$ and the anchor is fixed conditional on $\mathcal C$,
	\[
	\mathbb E[r_K(h^{\mathrm{aug}}_{\mathcal C}-h^0_{\mathcal C})\mid\mathcal C]\ge-T_n.
	\]
	Finally, $|h^{\mathrm{aug}}_{\mathcal C}|\le c_\alpha$ pointwise, so conditional Cauchy--Schwarz gives
	\[
	\mathbb E[(r_m-r_K)h^{\mathrm{aug}}_{\mathcal C}\mid\mathcal C]
	\ge-\|r_m-r_K\|_{L^2(\mathbb P_X)}
	\bigl(\mathbb E[(h^{\mathrm{aug}}_{\mathcal C})^2\mid\mathcal C]\bigr)^{1/2}
	\ge-c_\alpha\Delta_{K,m}.
	\]
	
	\medskip
	\noindent\emph{Step 6: Combine.}
	The left side of~\eqref{eq:projected-transfer} equals
	$\mathbb P(Y\in\widehat C_\alpha(X)\mid M=m,\mathcal C)-(1-\alpha)$.
	Combining its three bounds proves~\eqref{eq:mcov-thm} on the calibration-only event~\eqref{eq:cal-excess}.
	No finite-sample term introduces an explicit $1/\pi_m$ factor. Equivalently, using $\rho_K=\pi_mr_K$ would give coefficient norm at most $\pi_mK$, and the common factor $\pi_m$ would cancel at the final normalization. Dependence on mask rarity remains through $\Delta_{K,m}$.
\end{proof}

\subsection{Proof of \cref{thm:mask-cond-generalJ-revised} and the optimized rate}
\label{sec:rate-explicit}
\begin{proof}
	Here $n=u-1\ge u/2$. Let $L_\delta=\log(2/\delta)$, $a_0=2C_\ell^2B_\psi^2/\lambda_J$, and define constants
	\[
	E_0=\sqrt2\left(\frac{a_0}{\sqrt{\log2}}+\frac{4a_0+2B_\ell}{\sqrt2}\right),
	\quad L_0=C_0B_\psi^2+2\lambda_{J\mathrm{max}},
	\quad H_0=\bar C_0KB_\psi^2\sqrt{\frac{2B_\ell}{\lambda_J}}.
	\]
	For $0<\lambda\le1$, \eqref{eq:cal-excess-def} implies
	$\varepsilon_{\mathrm{cal}}(\delta)\le E_0L_\delta^{1/2}/(\lambda u^{1/2})$.
	Write $A=(1+\sqrt\lambda)L_\delta^{1/4}/(\sqrt\lambda u^{1/4})$. If $A<1$, then
	\[
	D_n+T_n\le
	\left(\frac{4K^2}{\kappa}\sqrt{2L_0E_0}
	+\frac{4K}{\kappa}E_0+\frac{H_0}{(\log2)^{1/4}}\right)A.
	\]
	We used $\varepsilon_{\mathrm{cal}}\le E_0A^2\le E_0A$ and the looser add-one bound for $T_n$. Choose $C_1$ as the maximum of $1$ and the coefficient in parentheses, and
	$C_2=\max\{1,2K\sqrt{\lambda_{J\mathrm{max}}B_\ell}\}$.
	When $A\ge1$ or $\lambda\ge1$, the desired lower bound is nonpositive and follows from nonnegativity of coverage. In the remaining case it follows from~\eqref{eq:mcov-thm}; $c_\alpha=1-\alpha$ for $\alpha\le1/2$.
\end{proof}

\paragraph{Uniform constants and the explicit rate.}
If the constants above are uniform over the chosen regularization values, minimizing
\[
C_1L_\delta^{1/4}u^{-1/4}(\lambda^{-1/2}+1)+C_2\lambda^{1/2}
\]
gives
\[
\lambda^\star=\frac{C_1}{C_2}L_\delta^{1/4}u^{-1/4},
\qquad
C_1L_\delta^{1/4}u^{-1/4}+2\sqrt{C_1C_2}L_\delta^{1/8}u^{-1/8}.
\]
Thus the leading finite-sample rate is $u^{-1/8}$ for fixed $\delta$, while the loose and sharp transfer terms are respectively $O(u^{-3/8})$ and $O(u^{-3/4})$ at this choice. The representation term is not asserted to vanish with $u$.

To control the constants uniformly in $\lambda$, we bound $\kappa$ away from zero. Under the bounded-score and Gram assumptions, a sufficient fixed-radius choice is available. Let $a_\alpha=\min\{\alpha,1-\alpha\}$ and $g_{\min}=\lambda_{\min}(G)$. Since $L(\beta^\star)\le L(0)$ and $\ell_\alpha(\theta,s)\ge a_\alpha(|\theta|-|s|)$,
\[
a_\alpha\mathbb E|g_{\beta^\star}(X)|\le\mathbb E|S|\le S_{\max}.
\]
Moreover, bounded features imply
$g_{\min}\|\beta^\star\|_2^2\le\mathbb E[g_{\beta^\star}(X)^2]
\le B_\psi\|\beta^\star\|_2\mathbb E|g_{\beta^\star}(X)|$.
Consequently
\[
\|\beta^\star\|_2\le R_{\mathrm{pop}}:=\frac{B_\psi S_{\max}}{a_\alpha g_{\min}}
\quad\hbox{for every }\lambda>0.
\]
Choosing a fixed $K>R_{\mathrm{pop}}$ yields the uniform margin $\kappa\ge K-R_{\mathrm{pop}}>0$ and hence the stated optimized rate for the bounded-class procedure, with no new distributional assumption. The constants $C_0,t_0$ are those assumed for this chosen ball. If $K$ or other constants instead vary with $u$, that dependence must be retained in the rate.

\paragraph{Relation to unconstrained optimization.}
For full-rank $J$, the unconstrained empirical minimizer satisfies a deterministic localization bound. The KKT condition and bounded pinball subgradients give
\[
2\lambda J^\top J\widehat\beta_{\mathcal D}
=-\frac1{|\mathcal D|}\sum_i b_i\psi(X^i),\quad |b_i|\le C_\ell,
\qquad
\|\widehat\beta_{\mathcal D}\|_2\le
R_{\mathrm{loc}}(\lambda):=\frac{C_\ell B_\psi}{2\lambda\lambda_J}.
\]
The same bound holds for the population minimizer and for every candidate score, even if scores are unbounded. Therefore a radius $K>R_{\mathrm{loc}}(\lambda)$ makes the constrained and unconstrained estimators coincide and ensures population interiority for that $\lambda$. Separately, the unrestricted projection coefficient is
\[
\eta_m^\star=G^{-1}\mathbb E[\psi(X)r_m(X)]
=G^{-1}\mathbb E[\psi(X)\mid M=m],
\qquad
\|\eta_m^\star\|_2\le B_\psi/g_{\min}.
\]
Thus $K>\max\{R_{\mathrm{loc}}(\lambda),B_\psi/g_{\min}\}$ suffices for both localization and projection feasibility. If $K$ is chosen as a function of $\lambda$, this dependence must also be included when interpreting the rate.

\paragraph{Unbounded scores by centering the loss.}
For unbounded scores, including those in the Gaussian experiments, we use a centered-loss extension. Assume $\mathbb E|S|<\infty$ and, on the same fixed $B_K$, define
\[
\bar\ell_\beta(z)=\ell_\alpha(g_\beta(x),s)-\ell_\alpha(0,s),
\qquad
|\bar\ell_\beta(z)|\le B_{\mathrm{cen}}:=C_\ell KB_\psi.
\]
The shifted loss $\bar\ell_\beta+B_{\mathrm{cen}}$ lies in $[0,2B_{\mathrm{cen}}]$ and has the same minimizers, gradient, loss-difference stability, and excess risk as the original loss. Applying the preceding proof with $B_\ell=2B_{\mathrm{cen}}$ gives~\eqref{eq:mcov-thm}; comparison with $\beta=0$ also directly gives $\lambda\|J\widetilde\beta_{\mathcal C}\|_2^2\le B_{\mathrm{cen}}$. The uniform population-radius argument above uses $\mathbb E|S|$ in place of $S_{\max}$.

For a CQR score $S=a(X)+|Y-c(X)|$ under independent additive $N(0,\sigma^2)$ noise, the conditional density of $Y$ given the masked covariates is a mixture of Gaussian densities bounded by $1/(\sqrt{2\pi}\sigma)$. The inverse image of a score window of radius $t$ has total length at most $4t$, hence
\[
\mathbb P(|S-q|\le t\mid X)\le\frac{4t}{\sqrt{2\pi}\sigma}.
\]

\subsection{Discussion of Remark \ref{remark:estimation}}
\begin{proposition}
	\label{prop:DeltaK-upper-r}
	Let 
	$\mu=\mathbb E[M]$ the mean and the covariance $\Sigma_M=\mathrm{Cov}(M)$, and fix a pattern $m\in\{0,1\}^d$ with $\pi=\mathbb P(M=m)>0$. 
	Consider the closed linear subspace 
	\[
	\mathcal H\;=\;\Big\{\,a_0+a^\top \mathsf m(x):~a_0\in\mathbb R,~a\in\mathbb R^d\,\Big\}
	,
	\]
	and let
	$
	r^\star(X) = a_0 + \sum_{j=1}^d a_j \mathsf{m}_j(X)$ 
	be the $L^2(\mathbb P)$-projection of $r_m$ onto $\mathcal H$; equivalently, 
	$(a_0^\star,a^\star)$ minimizes $\mathbb E\!\left[(r_m-a_0-a^\top M)^2\right]$. Then $
	a^\star=\Sigma_M^{\dagger}(m-\mu)$, 
	$a_0^\star=1-(a^\star)^\top\mu$,
	and the projection error is
	$
	\Delta_{\rm mask}^2
	~:=~{\frac{1-\pi}{\pi}}
	\;-\;
	(m-\mu)^\top \Sigma_M^{\dagger}(m-\mu).
	$
	Consequently, if the model class $\mathcal G_K$ used in Theorem~\ref{thm:mask-cond-generalJ-revised}
	contains its optimal projection, for example when $\|(a^\star,a_0^\star)\|_2\le K$, we have
	\[
	\Delta_{K,m}^2
	~=~\min_{g\in\mathcal G_K}\|r_m-g\|_{L^2(\mathbb P)}^2
	~\le~
	\Delta_{\rm mask}^2.
	\]
\end{proposition}

The proof of Proposition~\ref{prop:DeltaK-upper-r} follows from the normal equations for population least squares. Under the stated coefficient-feasibility condition, it supplies an upper bound on $\Delta_{K,m}$ in terms of $\pi$, $\mu$, and $\Sigma_M$. For an arbitrary radius $K$, the constrained mask-only class gives
\[
\Delta_{K,m}\le\inf_{h\in\mathcal H_K^{\mathrm{mask}}}\|r_m-h\|_{L^2(\mathbb P_X)},\ 
\mathcal H_K^{\mathrm{mask}}=\{a_0+a^\top\mathsf m(\cdot):\|(a,a_0)\|_2\le K\}\subseteq\mathcal G_K.
\]
In the special case of independent mask coordinates, the bound admits a closed-form expression under the same coefficient-feasibility condition.

\begin{proposition}[Explicit projection error with mask + tree features]
	\label{prop:DeltaK-upper-mask-tree}
	Fix a pattern $m\in\{0,1\}^d$ with $\pi=\mathbb P(M=m)>0$ and define
	\[
	r_m(X):=\frac{\ind\{M=m\}}{\pi}.
	\]
	Let the tree embedding be $\widehat T(X)\in\mathbb R^{n_{\tree}}$ and define the joint feature vector
	\[
	\chi(X):=
	\begin{bmatrix}
		\widehat T(X)\\
		M
	\end{bmatrix}
	\in \mathbb R^{d+n_{\tree}}.
	\]
	Consider the closed linear space
	\[
	\mathcal H_{\mathrm{mask+tree}}
	=
	\Big\{\,h_{\beta_0,\beta}(x)=\beta_0+\beta^\top \chi(x):~\beta_0\in\mathbb R,\;\beta\in\mathbb R^{d+n_{\tree}}\,\Big\}.
	\]
	Let $(\beta_0^\star,\beta^\star)$ be the $L^2(\mathbb P)$-projection of $r_m$ onto $\mathcal H_{\mathrm{mask+tree}}$, i.e.
	\[
	(\beta_0^\star,\beta^\star)\in\arg\min_{\beta_0,\beta}\;
	\mathbb E\Big[(r_m(X)-\beta_0-\beta^\top \chi(X))^2\Big].
	\]
	Write $\mu_\chi=\mathbb E[\chi]$ and $\Sigma_\chi=\mathrm{Cov}(\chi)$.
	Then we have
	\[
	\beta^\star=\Sigma_\chi^{\dagger}\,v_m, 
	\beta_0^\star=1-(\beta^\star)^\top \mu_\chi,
	\]
	where
	\[
	v_m:=\mathrm{Cov}(\chi,r_m)=\mathbb E[\chi\mid M=m]-\mathbb E[\chi]
	=
	\begin{bmatrix}
		\mathbb E[\widehat T(X)\mid M=m]-\mathbb E[\widehat T(X)]\\
		m-\mu
	\end{bmatrix},
	\quad
	\mu=\mathbb E[M].
	\]
	Moreover, the corresponding squared projection error admits the explicit form
	\[
	\Delta_{\mathrm{mask+tree}}^2
	:=
	\min_{h\in\mathcal H_{\mathrm{mask+tree}}}
	\|r_m-h\|_{L^2(\mathbb P)}^2
	=
	\frac{1-\pi}{\pi}
	-
	v_m^\top \Sigma_\chi^{\dagger} v_m.
	\]
	In particular, since $\mathcal H\subseteq\mathcal H_{\mathrm{mask+tree}}$, $\Delta_{\mathrm{mask+tree}}^2\le\Delta_{\mathrm{mask}}^2$ for the unrestricted spaces. The equality $\Delta_{K,m}=\Delta_{\mathrm{mask+tree}}$ holds if an unrestricted optimal coefficient vector $(\beta^\star,\beta_0^\star)$ has norm at most $K$. Otherwise $\Delta_{K,m}\ge\Delta_{\mathrm{mask+tree}}$, and the finite-radius comparison must use the constrained projection.
\end{proposition}

This is a direct consequence of the normal equations for population least squares.
Adding tree coordinates cannot increase the approximation error when the bounded mask-only class is embedded by zero padding in the same-radius joint class. Strict improvement requires the residualized tree features to explain a nonzero component of the mask indicator; a tree feature duplicating a mask coordinate gives no improvement. If $r_m$ is exactly representable by a feasible coefficient vector in $\mathcal G_K$, then $\Delta_{K,m}=0$; approximate representation gives a correspondingly small residual.

\subsection{Proof of Proposition~\ref{prop:strong-duality}}
\label{proof:strong duality}

\begin{proof}
	 Multiplying \eqref{eq:regularised loss} by
	$u=|\mathcal I_2|+1$ and introducing the standard pinball-loss slack variables
	$a_i,b_i\ge0$ yields a finite-dimensional convex program with equality
	constraints
	\[
	S^i-g_\beta(X^i)-a_i+b_i=0,\quad i\in\mathcal I_2,
	\qquad
	s-g_\beta(X^{n+1})-a_{n+1}+b_{n+1}=0.
	\]
	Since $J^\top J\succeq0$, the objective is convex. Slater's condition holds
	because the slack variables can be chosen strictly positive while satisfying the
	equalities. Therefore, strong duality holds. Applying the same Lagrangian calculation as in \citet{gibbs2025conformal},
	minimization over the slack variables gives
	$-\alpha\le\eta_i\le1-\alpha$, and minimization over $\beta$ gives exactly the
	sample-dependent conjugate
	$\mathcal R_{\mathcal I_2,n+1}^\star(\eta)$ defined in
	Proposition~\ref{prop:strong-duality}.  For the bounded class, $\beta=0$ together with sufficiently large positive slack variables is strictly feasible when $K>0$, and the same derivation restricts the conjugate minimization to $B_K$.
\end{proof}

While the original dual-optimization problem is computationally feasible, those steps require several root-finding procedures for each test point.
A natural plug-in approximation uses $\widetilde g=g_{\widetilde\beta_{\mathcal C}}$, the calibration-only fit with normalization $1/n$, where $n=|\mathcal I_2|$, instead of the candidate-augmented fit with normalization $1/(n+1)$. Proposition~\ref{prop:plugin} controls this replacement in the full-rank regularized regime.

\begin{proposition}[Plug-in approximation]\label{prop:plugin}
	Under the assumptions of \cref{thm:mask-cond-generalJ-revised}, with both fits over the same $B_K$, let $u=n+1$. Uniformly over $x$ and candidate pairs in the score support,
	\[
	|g_{\widehat\beta_{\mathcal C,z}}(x)-\widetilde g(x)|
	\le\tau_n
	\le\frac{C}{\sqrt u},\ 
	C=B_\psi\sqrt{\frac{2B_\ell}{\lambda\lambda_J}},
	\]
	where $\tau_n$ is defined in~\eqref{eq:repair-constants}. This is an $O(u^{-1/2})$ bound for fixed $\lambda$ and fixed constants, with the sharper $O((\lambda u)^{-1})$ bound also available.
	If the conditional CDF of $S$ given $X=x$ is uniformly $L_S$-Lipschitz, then
	\[
	\left|\mathbb P(S\le g_{\widehat\beta_{\mathcal C,Z}}(X)\mid X=x,\mathcal C)
	-\mathbb P(S\le\widetilde g(X)\mid X=x,\mathcal C)\right|
	\le\min\{1,2L_S\tau_n\}.
	\]
	The same bound then holds after averaging over $X$ conditional on $M=m$.
\end{proposition}

\subsection{Proof of Proposition~\ref{prop:plugin}}
\begin{proof}
	Equation~\eqref{eq:add-one} proves the threshold bound with strong-convexity constant $\mu=2\lambda\lambda_J$.
	The two inclusion indicators differ only within $\tau_n$ of the calibration-only threshold. Conditional on $X=x$ and $\mathcal C$, that threshold is fixed, so the Lipschitz CDF bounds the window probability by $2L_S\tau_n$.
\end{proof}

\section{Additional experiments}
\subsection{Experimental details and reproducibility}
\label{sec:exp-details}
This section reports the sample splits and implementation choices used to generate the main experimental figures.
All experiments use $1-\alpha=90\%$ and 10 random replications.
The base lower and upper quantile models are HistGBDT models tuned with 100 Optuna trials using only the selection half of the proper training pool; all feature transformations and $\lambda$ selection are likewise fixed before calibration. The base-fit half supplies model fitting, and the calibration sample is used only for the final conformal step. Table~\ref{tab:clean-reruns} compares this training-only protocol with tuning the base models and $\lambda$ on calibration data, illustrating sensitivity to the tuning split.

\begin{table}[ht]
	\centering
	\caption{Data splits used in the main experiments. The reported mask-wise metrics are computed on the pattern-stratified test set.}
	\label{tab:exp-splits}
	\begin{tabular}{lcccc}
		\toprule
		Experiment & $d$ & Training pool & Calibration & Pattern-stratified test set \\
		\midrule
		Synthetic, low-dimensional & 3 and 6 & 1000 & 500 & $100\times|\mathcal M_{\mathrm{test}}|$ \\
		Synthetic, high-dimensional & 100 & 5000 & 1000 & $100\times100=10000$ \\
		Concrete & 8 & 630 & 200 & $100\times100=10000$ \\
		MIMIC-IV LOS & 300 & 3000 & 1000 & 2331 \\
		\bottomrule
	\end{tabular}
\end{table}

For the low-dimensional synthetic experiments, the pattern-stratified test set contains 100 independent test samples for each allowed missingness mask.
Thus, under MCAR and PMIX it contains 800 points for $d=3$ and 6400 points for $d=6$; under MAR, only the first missingness-active coordinates are varied, giving 400 points for $d=3$ and 800 points for $d=6$.
The high-dimensional synthetic experiment uses $d=100$, a training pool of 5000 points, 1000 calibration points, and 100 randomly sampled test masks with 100 test points per mask.
For the Concrete benchmark, the 630 training points are followed by 200 calibration points and a pattern-stratified test set of 100 masks with 100 points per mask.
For MIMIC-IV LOS, each replication samples 3000 training rows and 1000 calibration rows from the official training file; the fixed test file contains 2331 rows spanning 50 missingness patterns.

For reproducibility, the base quantile models are tuned over learning rate in $[10^{-3},0.2]$ on a log scale, max iterations in $\{50,\ldots,400\}$, max depth in $\{2,\ldots,6\}$, minimum samples per leaf in $\{10,\ldots,80\}$, $\ell_2$ regularization in $[10^{-6},10]$ on a log scale, and max bins in $\{128,\ldots,255\}$, using a TPE sampler.
For the tree-embedding model, the number of boosting rounds is fixed to the reported embedding dimension $n_{\tree}$, and the regularization parameter $\lambda$ in the model is selected from 50 log-spaced values in $[10^{-4},10^3]$ by repeated 5-fold cross-validation on the training-side selection half, never on calibration data. For the MICE baseline, the imputer is fitted using proper training data only, and applied without refitting to calibration and test inputs.

\paragraph{Sensitivity to the tuning protocol.}
Table~\ref{tab:clean-reruns} compares calibration-based and training-only tuning. Under training-only tuning, the synthetic results retain coverage above the nominal level with smaller shortfall and somewhat longer intervals. On MIMIC, coverage remains above $0.9$, while shortfall and interval length increase.
\begin{table}[htbp]
	\centering
	\caption{Sensitivity to the tuning protocol (10-seed means). Each entry is calibration-based tuning $\to$ training-only tuning. Synthetic settings use $d=6$, $p=0.2$, and nominal coverage $0.9$.}
	\label{tab:clean-reruns}
	\begin{tabular}{lccc}
		\toprule
		Setting & Coverage & Shortfall & Interval length\\
		\midrule
		MCAR & $0.9013\to0.9162$ & $0.0062\to0.0031$ & $7.505\to8.120$\\
		MAR & $0.9020\to0.9120$ & $0.0013\to0.0002$ & $5.776\to6.008$\\
		PMIX & $0.9068\to0.9145$ & $0.0234\to0.0131$ & $9.728\to10.917$\\
		MIMIC-IV & $0.9278\to0.9225$ & $0.0090\to0.0152$ & $107.882\to114.814$\\
		\bottomrule
	\end{tabular}
\end{table}

\paragraph{Oracle length.}
The ``Oracle Length'' curve is a population benchmark for the intrinsic oracle $90\%$ conditional interval length of $Y$ given the observed coordinates and the mask.
In the Gaussian linear MCAR model, for a mask $m$ with missing coordinates $\mathrm{mis}(m)$ and observed coordinates $\mathrm{obs}(m)$, this length is
\[
2z_{1-\alpha/2}
\left(
1+
\beta_{\mathrm{mis}}^\top
\Sigma_{\mathrm{mis}\mid\mathrm{obs}}
\beta_{\mathrm{mis}}
\right)^{1/2},
\]
where $\Sigma_{\mathrm{mis}\mid\mathrm{obs}}$ is the conditional covariance of the missing coordinates given the observed coordinates.
For MAR/MNAR synthetic settings, we use Monte Carlo to approximate this length. We do not use it as a (semi-)real data baseline, since the oracle distribution is unavailable for Concrete and MIMIC-IV.

\begin{figure}[ht]
	\centering
	\includegraphics[width=0.9\linewidth]{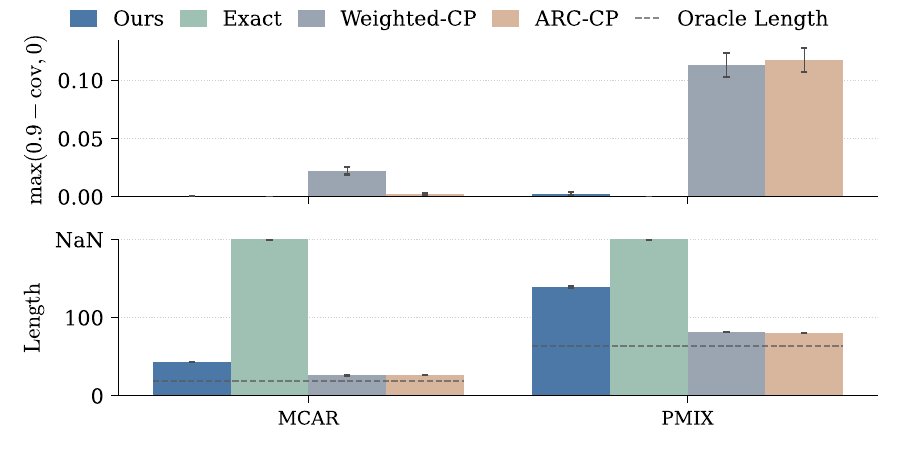}
	\caption{
			Empirical shortfall risk and average interval length for the high-dimensional synthetic experiment ($d=100$, $p=0.2$) under MCAR and PMIX mechanisms. Our method keeps shortfall risk near zero for both mechanisms while maintaining informative intervals.
		}
	\label{fig:dim100-shortfall}
\end{figure}

\paragraph{Exact dual versus plug-in.}
All coverage and length figures use the exact dual computation: low-dimensional synthetic Figures~\ref{fig:shortfall}--\ref{fig:dim6_ntree}, the high-dimensional synthetic result in Figure~\ref{fig:dim100-shortfall}, the Concrete result in Figure~\ref{fig:concrete-shortfall}, and the MIMIC-IV result in Figure~\ref{fig:mimic-combined}.
\subsection{The cross-validation of lambda}
\begin{figure}[ht]
	\centering
	\includegraphics[width=0.8\linewidth]{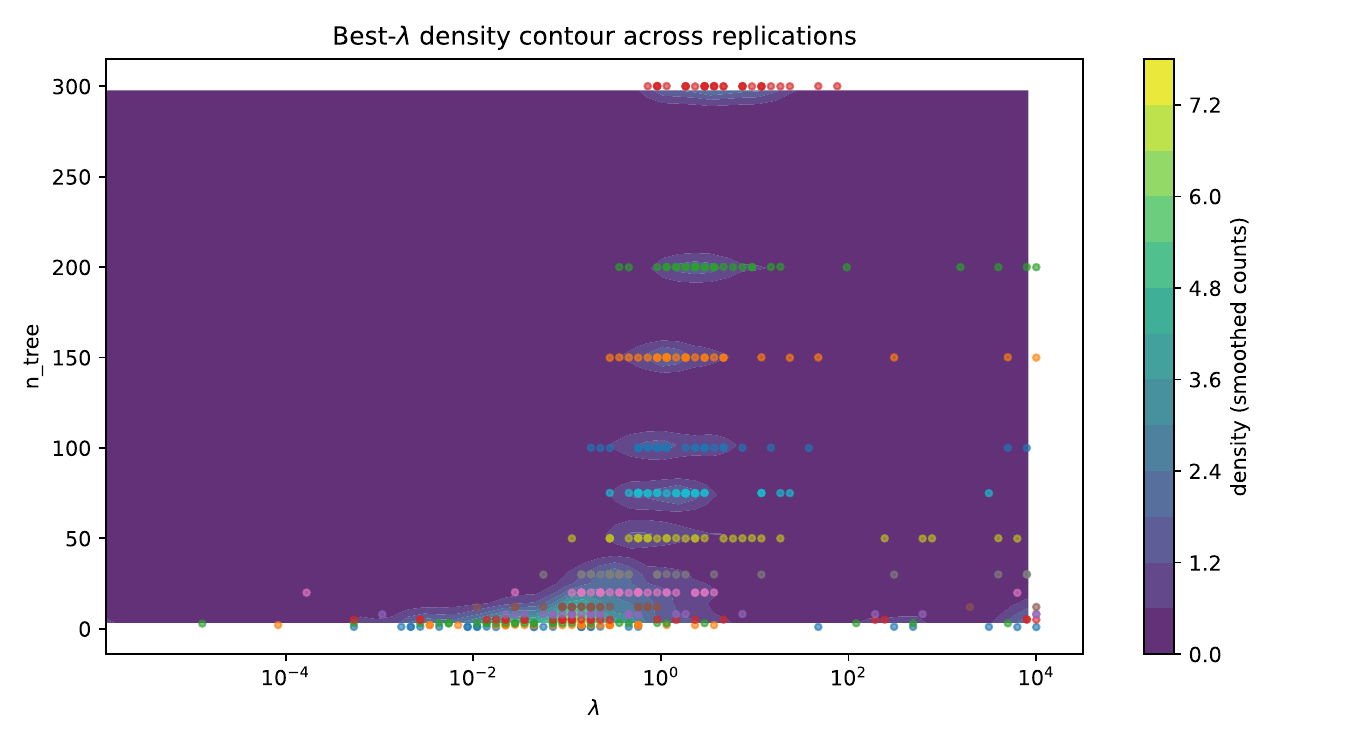}
	\caption{
		Choice of the best $\lambda$ over different numbers of trees and replications
	}
	\label{fig:lambda-surface}
\end{figure}

Our quantile estimator $g_\beta$ is fitted by minimizing the regularized
pinball loss in~\eqref{eq:minimizer}, which involves a ridge-type penalty
controlled by $\lambda$. Figure~\ref{fig:lambda-surface} shows the distribution of selected $\lambda$
values as a function of the embedding dimension (number of trees $n_{\tree}$)
on the synthetic $d=6$ MCAR setting. Each dot corresponds to one replication,
and the contour plot visualizes the empirical density of the best $\lambda$. We observe that the optimal $\lambda$ lies within a relatively narrow range
across a wide range of $n_{\tree}$, suggesting that in practice it is sufficient to choose the best $\lambda$ from a small set of candidates.

\subsection{Runtime}
\label{sec:runtime}
We report empirical runtime to clarify the computational trade-off between the exact dual path and the plug-in approximation. All timings are measured on an Intel Xeon Gold 6252 CPU @ 2.10GHz, with each job limited to 8 processes. Table~\ref{tab:runtime-methods} reports average time per test point. In low dimensions, the exact version of our method is slower than the cheapest baselines, while the plug-in approximation is much closer to them. In the high-dimensional setting, MDA-Nested does not finish within a reasonable time, whereas our method remains feasible.

\begin{table}[ht]
	\centering
	\caption{Runtime per test point in seconds.}
	\label{tab:runtime-methods}
	\begin{tabular}{lcc}
		\toprule
		Method & $d=6$ & $d=100$ \\
		\midrule
		ARC-CP & 0.0054 & 0.0079 \\
		MDA-Exact & 0.0057 & 0.0073 \\
		Ours (the exact dual optimization) & 1.0594 & 2.2980 \\
		Weighted CP & 0.2316 & 0.4034 \\
		MDA-Nested & 0.8275 & Did not finish \\
		\bottomrule
	\end{tabular}
\end{table}

Table~\ref{tab:runtime-scaling} shows how the exact computation changes with calibration size and embedding dimension on the $d=6$ synthetic setting. The cost increases with both quantities, consistent with the dual optimization problem, depending on the calibration sample size and the dimension of the learned representation.

\begin{table}[ht]
	\centering
	\caption{Runtime scaling for our exact method on the $d=6$ synthetic setting.}
	\label{tab:runtime-scaling}
	\begin{tabular}{lccccc}
		\toprule
		Calibration size & 100 & 200 & 500 & 1000 & 2000 \\
		\midrule
		Time (s) & 0.2454 & 0.3585 & 0.7130 & 1.0594 & 2.0533 \\
		\bottomrule
	\end{tabular}
	
	\vspace{0.8em}
	\begin{tabular}{lccccc}
		\toprule
		$n_{\tree}$ & 4 & 16 & 64 & 256 & 512 \\
		\midrule
		Time (s) & 0.2393 & 0.2982 & 0.4242 & 1.0594 & 1.7268 \\
		\bottomrule
	\end{tabular}
\end{table}

\subsection{Observed vs.\ Unobserved Mask Patterns}

Our method attains near-zero shortfall risk across all missingness mechanisms, but uses slightly wider intervals than the baselines. A natural question is whether this conservativeness is uniform across masks or arises primarily from mask patterns that are not represented in the calibration set. Here we compare the performance on the PMIX synthetic dataset for 2 subgroups of the test set.

Let $\mathcal M_{\mathrm{cal}}$ denote the set of missingness masks that appear in the calibration sample. We compare performance on
(i) $m \in \mathcal M_{\mathrm{cal}}$ (``observed masks''), and
(ii) $m \notin \mathcal M_{\mathrm{cal}}$ (``unobserved masks'').

\begin{figure}[ht]
	\centering
	\includegraphics[width=0.95\linewidth]{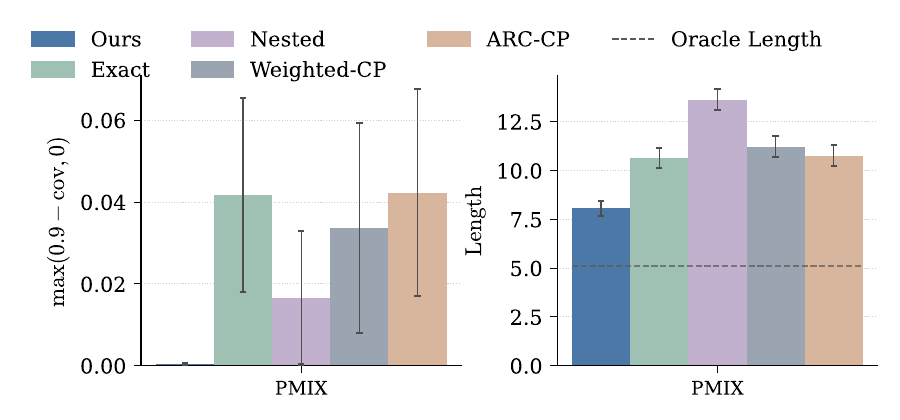}
	\caption{ Risk and average interval length for all $m \in \mathcal{M}_{cal}$. For the mask observed in calibration data, we have a near-zero risk for miscoverage, and the average interval length is the smallest.
	}
	\label{fig:d6_pmix_seenmask}
\end{figure}

\begin{figure}[ht]
	\centering
	\includegraphics[width=0.95\linewidth]{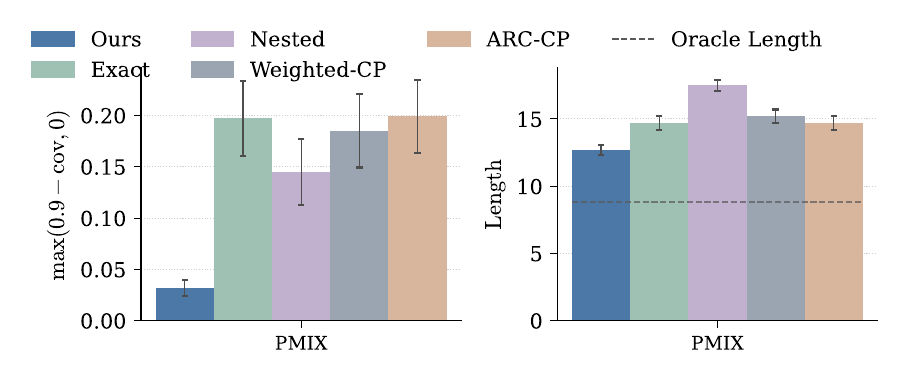}
	\caption{ Risk and average interval length for all $m \notin \mathcal{M}_{cal}$. For the mask not observed in calibration data, the risk is slightly bigger, and the interval length is also largely augmented, although it is still the shortest interval, showing that our method can be adapted to "easy" masks and "difficult" masks.
	}
	\label{fig:d6_pmix_unseenmask}
\end{figure}

Figure~\ref{fig:d6_pmix_seenmask} shows that for $m \in \mathcal M_{\mathrm{cal}}$ our method achieves near-zero shortfall risk with the \emph{smallest} average interval length among all competitors. By contrast, for $m \notin \mathcal M_{\mathrm{cal}}$ (Figure~\ref{fig:d6_pmix_unseenmask}) the shortfall risk is slightly higher and intervals are wider, while still remaining shorter than those of the baselines. This suggests that our procedure automatically adapts: it produces tight intervals for ``easy'' (data-supported) mask patterns and becomes more conservative only for ``difficult'' (data-poor) masks.

\subsection{The choice of embedding model}
\begin{figure}[ht]
	\centering
	\includegraphics[width=0.8\linewidth]{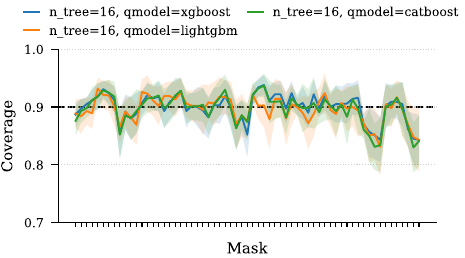}
	\caption{
		Empirical coverage over different embedding tree models.
	}
	\label{fig:embed-model-cov}
\end{figure}

\begin{figure}[ht]
	\centering
	\includegraphics[width=0.8\linewidth]{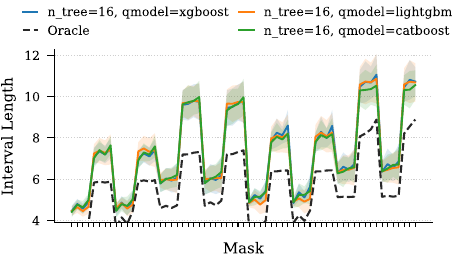}
	\caption{
		Empirical interval length over different embedding tree models.
	}
	\label{fig:embed-model-len}
\end{figure}

Our method only requires a tree ensemble $f_{\text{tree}}$ to define the embedding
$\widehat T(x)$; in principle, any off-the-shelf tree-based quantile regressor or other models
could be used. In all main experiments we instantiate $f_{\text{tree}}$ with
\texttt{LightGBM}, as it supports missing values natively and scales well to
high-dimensional data. To assess the sensitivity to this design choice, we compare several alternative
embedding models, including XGBoost and CatBoost. Figure~\ref{fig:embed-model-cov} reports the marginal coverage and mask-conditional
coverage for each embedding model, while Figure~\ref{fig:embed-model-len} shows
the corresponding interval lengths

\subsection{High dimension experiments}
%

\clearpage
\newpage
\input{checklist.tex}

\end{document}

%% file: checklist.tex
\section*{NeurIPS Paper Checklist}

\begin{enumerate}

\item {\bf Claims}
    \item[] Question: Do the main claims made in the abstract and introduction accurately reflect the paper's contributions and scope?
    \item[] Answer: \answerYes{} 
    \item[] Justification: The main claims in abstract and introduction correspond to our contributions.

\item {\bf Limitations}
    \item[] Question: Does the paper discuss the limitations of the work performed by the authors?
    \item[] Answer: \answerYes{} 
    \item[] Justification: We include our limitation in the conclusion section.

\item {\bf Theory assumptions and proofs}
    \item[] Question: For each theoretical result, does the paper provide the full set of assumptions and a complete (and correct) proof?
    \item[] Answer: \answerYes{} 
    \item[] Justification: We include assumptions and proofs for each theorem and proposition.{}

    \item {\bf Experimental result reproducibility}
    \item[] Question: Does the paper fully disclose all the information needed to reproduce the main experimental results of the paper to the extent that it affects the main claims and/or conclusions of the paper (regardless of whether the code and data are provided or not)?
    \item[] Answer: \answerYes{} 
    \item[] Justification: We provide the dataset, missingness mechanism and the base models. We provide also different augmentations and penalty to be used in each experiment.{}

\item {\bf Open access to data and code}
    \item[] Question: Does the paper provide open access to the data and code, with sufficient instructions to faithfully reproduce the main experimental results, as described in supplemental material?
    \item[] Answer: \answerYes{} 
    \item[] Justification: We provide all code to reproduce the experiments, except the dataset MIMIC-IV, which must be requested on physionet. {}

\item {\bf Experimental setting/details}
    \item[] Question: Does the paper specify all the training and test details (e.g., data splits, hyperparameters, how they were chosen, type of optimizer) necessary to understand the results?
    \item[] Answer: \answerYes{} 
    \item[] Justification: For each dataset, we present the split number of training/calibration/test set.{}

\item {\bf Experiment statistical significance}
    \item[] Question: Does the paper report error bars suitably and correctly defined or other appropriate information about the statistical significance of the experiments?
    \item[] Answer: \answerYes{} 
    \item[] Justification: For each experiment, we do 10 replications to show the statistical confidence interval for miscoverage and interval length.

\item {\bf Experiments compute resources}
    \item[] Question: For each experiment, does the paper provide sufficient information on the computer resources (type of compute workers, memory, time of execution) needed to reproduce the experiments?
    \item[] Answer: \answerYes{} 
    \item[] Justification: We provide in the appendix a thorough running time comparison as well as compute resources.
    
\item {\bf Code of ethics}
    \item[] Question: Does the research conducted in the paper conform, in every respect, with the NeurIPS Code of Ethics \url{https://neurips.cc/public/EthicsGuidelines}?
    \item[] Answer: \answerYes{} 
    \item[] Justification: We follow the NeurIPS Code of Ethics.

\item {\bf Broader impacts}
    \item[] Question: Does the paper discuss both potential positive societal impacts and negative societal impacts of the work performed?
    \item[] Answer: \answerYes{} 
    \item[] Justification: The work studies uncertainty quantification under missing values, including an ICU length-of-stay benchmark. Potential positive impacts include better calibrated uncertainty in settings with incomplete measurements. Potential negative impacts include over-reliance on uncertainty intervals in high-stakes domains if assumptions, representation quality, or calibration data are inadequate.
    
\item {\bf Safeguards}
    \item[] Question: Does the paper describe safeguards that have been put in place for responsible release of data or models that have a high risk for misuse (e.g., pre-trained language models, image generators, or scraped datasets)?
    \item[] Answer: \answerNA{} 
    \item[] Justification: the paper poses no such risks.

\item {\bf Licenses for existing assets}
    \item[] Question: Are the creators or original owners of assets (e.g., code, data, models), used in the paper, properly credited and are the license and terms of use explicitly mentioned and properly respected?
    \item[] Answer: \answerYes{} 
    \item[] Justification: We cited the UCI Concrete dataset and MIMIC-IV dataset that are used in the paper.

\item {\bf New assets}
    \item[] Question: Are new assets introduced in the paper well documented and is the documentation provided alongside the assets?
    \item[] Answer: \answerNA{} 
    \item[] Justification: the paper does not release new assets.

\item {\bf Crowdsourcing and research with human subjects}
    \item[] Question: For crowdsourcing experiments and research with human subjects, does the paper include the full text of instructions given to participants and screenshots, if applicable, as well as details about compensation (if any)? 
    \item[] Answer: \answerNA{} 
    \item[] Justification: the paper does not involve crowdsourcing nor research with human subjects.

\item {\bf Institutional review board (IRB) approvals or equivalent for research with human subjects}
    \item[] Question: Does the paper describe potential risks incurred by study participants, whether such risks were disclosed to the subjects, and whether Institutional Review Board (IRB) approvals (or an equivalent approval/review based on the requirements of your country or institution) were obtained?
    \item[] Answer: \answerNA{} 
    \item[] Justification: the paper does not involve crowdsourcing nor research with human subjects.

\item {\bf Declaration of LLM usage}
    \item[] Question: Does the paper describe the usage of LLMs if it is an important, original, or non-standard component of the core methods in this research? Note that if the LLM is used only for writing, editing, or formatting purposes and does \emph{not} impact the core methodology, scientific rigor, or originality of the research, declaration is not required.
    \item[] Answer: \answerNA{} 
    \item[] Justification: the core method development in this research does not involve LLMs as any important, original, or non-standard components.

\end{enumerate}